\documentclass[letterpaper]{article} 

\usepackage{aaai2026}  
\usepackage{times}  
\usepackage{helvet}  
\usepackage{courier}  
\usepackage[hyphens]{url}  
\usepackage{graphicx} 
\usepackage{natbib}  
\usepackage{caption} 
\usepackage{amssymb}
\usepackage{amsmath}
\usepackage{booktabs}
\usepackage{multirow}
\usepackage{algorithm}
\usepackage{algorithmic}

\newcommand{\Lspace}{\mathbb{L}^n}
\newcommand{\Rspace}{\mathbb{R}^n}
\newcommand{\norm}[1]{\Vert #1 \Vert}

\usepackage{newfloat}
\usepackage{listings}
\DeclareCaptionStyle{ruled}{labelfont=normalfont,labelsep=colon,strut=off} 
\floatstyle{ruled}
\newfloat{listing}{tb}{lst}{}
\floatname{listing}{Listing}
\title{HyFI: Hyperbolic Feature Interpolation for Brain-Vision Alignment}

\author{
    Sangmin Jo,
    Wootaek Jeong,
    Da-Woon Heo,
    Yoohwan Hwang,
    Heung-Il Suk\thanks{Corresponding author}
}

\affiliations{
    Department of Artificial Intelligence, Korea University, Seoul, South Korea\\
    \text{\{sangminjo, wtjeong, daheo, yoohwan98, hisuk\}@korea.ac.kr}
}

\usepackage{bibentry}

\begin{document}

\maketitle

\begin{abstract}
Recent progress in artificial intelligence has encouraged numerous attempts to understand and decode human visual system from brain signals. These prior works typically align neural activity independently with semantic and perceptual features extracted from images using pre-trained vision models. However, they fail to account for two key challenges: (1) the modality gap arising from the natural difference in the information level of representation between brain signals and images, and (2) the fact that semantic and perceptual features are highly entangled within neural activity. To address these issues, we utilize hyperbolic space, which is well-suited for considering differences in the amount of information and has the geometric property that geodesics between two points naturally bend toward the origin, where the representational capacity is lower. Leveraging these properties, we propose a novel framework, \textbf{Hyperbolic Feature Interpolation (HyFI)}, which interpolates between semantic and perceptual visual features along hyperbolic geodesics. This enables both the fusion and compression of perceptual and semantic information, effectively reflecting the limited expressiveness of brain signals and the entangled nature of these features. As a result, it facilitates better alignment between brain and visual features. We demonstrate that HyFI achieves state-of-the-art performance in zero-shot brain-to-image retrieval, outperforming prior methods with Top-1 accuracy improvements of up to +$17.3\%$ on THINGS-EEG and +$9.1\%$ on THINGS-MEG. 
\end{abstract}

\begin{links}
\link{Code}{https://github.com/ku-milab/HyFI}
\end{links}

\section{Introduction}
Understanding how the human brain encodes information has long been a central topic in neuroscience, and has recently attracted growing attention in artificial intelligence. Specifically, the field of brain decoding aims to infer internal cognitive states or external sensory experiences from recorded brain activity \cite{naselaris2011encoding, oota2023deep}. It offers insights into how the brain represents the external world and enables brain-computer interface (BCI) systems \cite{ko2021multi}. In recent years, brain decoding techniques based on machine learning have been successfully applied across diverse cognitive domains such as vision, audition, and language \cite{wang2022open, defossez2023decoding, scotti2024mindeye2}. Among these, visual brain decoding has received particular attention, given that vision is the dominant sensory modality in humans and plays a crucial role in perception and cognition \cite{mathis2024decoding}.

\begin{figure}[t!]
    \centering
    \includegraphics[width=1\linewidth]{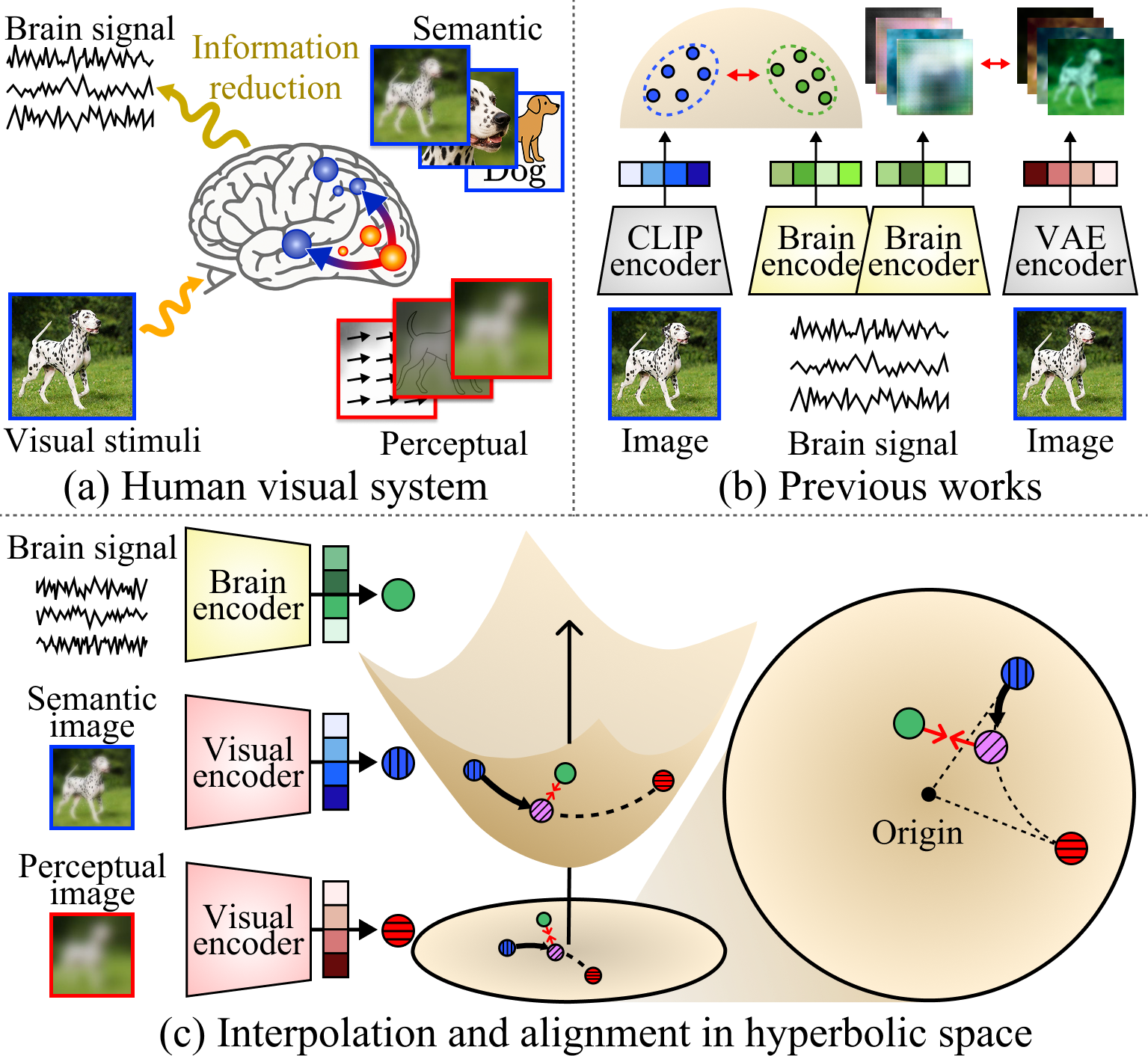}
    \caption{(a) The human visual system processes perceptual and semantic information, and some degradation occurs when neural activity is recorded. (b) Previous works aligned semantic and perceptual features through separate pathways, overlooking their entanglement in brain signals. (c) In contrast, hyperbolic interpolation merges perceptual and semantic features with lower complexity, enhancing alignment with brain signals.
 }
    \label{intro}
\end{figure}

Visual brain decoding has been extensively studied using neuroimaging modalities such as functional magnetic resonance imaging (fMRI), electroencephalography (EEG), and magnetoencephalography (MEG). In particular, fMRI has long been a dominant modality in this field due to its superior spatial resolution. However, its limited temporal resolution and bulky equipment make it less suitable for real-world applications. In contrast, EEG and MEG offer high temporal resolution, making it particularly suitable for BCI. Motivated by these advantages, we focus on brain decoding tasks based on EEG and MEG signals.

Recent brain decoding studies have increasingly adopted dual-pathway frameworks to capture both perceptual details and semantic representations from neural signals. This approach reflects the human visual system, which processes both \textit{perceptual} and \textit{semantic} features, as shown in Fig.~\ref{intro}(a). Perceptual features refer to low-level visual attributes extracted in early visual areas (e.g., V1), such as orientation, color, and edge information \cite{miyawaki2008visual}. Semantic features denote high-level conceptual representations encoded in cortical regions, such as object identity and category \cite{dicarlo2012does}. In recent approaches, semantic features are typically captured by aligning brain activity with image embeddings from pre-trained vision-language models (VLMs) like CLIP \citep{scotti2023reconstructing}. In parallel, perceptual features are often decoded by aligning representations derived from variational autoencoders (VAEs) \cite{shen2025neuro}. These efforts have advanced brain decoding by integrating brain-inspired models with multi-modal representations \citep{li2025neuraldiffuser}.

Despite these advances, current approaches still face two key limitations. First, aligning brain signals with image embeddings remains challenging—a problem commonly referred to as the modality gap \cite{liang2022mind}. This issue is known to arise from an inherent information imbalance between modalities \citep{,schrodi2024two}. In brain decoding, this imbalance is particularly pronounced, as neural signals contain substantially less semantic information than image embeddings for visual tasks \citep{chen2024visual, wu2025bridging}. The limitation is driven by human attentional bottlenecks, restricted visual working memory \citep{cavanagh2005tracking, dux2009attentional}, and the low signal-to-noise ratio and resolution of neural recordings \citep{srinivasan2007eeg, naselaris2011encoding}.
Second, many existing approaches that separately align perceptual and semantic features fail to reflect the fact that neural activity encodes features in a highly entangled and interactive manner, as they are not processed independently \cite{pollen1999neural, naselaris2009bayesian}. As a result, attempts to align these features independently may lead to suboptimal performance.

To address these limitations, we adopt hyperbolic space for brain-vision alignment. Unlike Euclidean space, hyperbolic geometry with negative curvature offers two key advantages: (1) geodesics between two points naturally bend toward the origin, (2) representational capacity decreases near the origin due to the exponential expansion of the space with radius. Building on these insights, we introduce a novel \textbf{Hy}perbolic \textbf{F}eature \textbf{I}nterpolation \textbf{(HyFI)} method that interpolates semantic and perceptual visual features in hyperbolic space. This allows semantic and perceptual features to be effectively integrated and compressed during interpolation, making it well suited for brain signals with limited information and entangled semantic-perceptual components. As a result, the interpolated representations become better aligned with brain activity. Our main contributions are as follows:

\begin{itemize}
    \item We propose a hyperbolic interpolation method that effectively integrates and compresses semantic and perceptual visual features, explicitly accounting for the limited information capacity and entangled nature of brain signals.

    \item Our method consistently improves performance across combinations of visual and brain encoders, demonstrating broad applicability.

    \item It achieves state-of-the-art (SOTA) performance on two public brain decoding benchmarks, with 68.2$\%$ Top-1 accuracy on THINGS-EEG and 35.8$\%$ on THINGS-MEG, outperforming previous methods by +17.3$\%$ and +9.1$\%$, respectively.

\end{itemize}

\section{Related Works}
\subsection{Visual Brain Decoding}
Visual brain decoding has received increasing attention for its potential to uncover the mechanisms of human cognition and to enable practical BCI \cite{kay2008identifying, yang2024brain}. With the rise of large-scale VLMs, recent studies have begun to leverage those representations by aligning semantic features with CLIP \cite{scotti2023reconstructing, songdecoding}. 
A similar trend is observed in EEG-based visual decoding. However, the inherent modality gap between neural signals and pre-trained visual embeddings remains a major challenge. To mitigate this, \citet{li2024visual} utilized a diffusion prior to map brain features into image space, \citet{zhang2025cognitioncapturer} employed multi-modal fusion to increase shared information, and \citet{wu2025bridging} reduced visual complexity using Gaussian blur.
These methods overlook the fact that semantic and perceptual representations are often entangled in neural activity. In contrast, our approach uses hyperboloid interpolation to fuse both features while reducing complexity, enabling better alignment.

\subsection{Hyperbolic Representation Learning} Hyperbolic space has attracted attention in representation learning for its ability to model hierarchical data due to its negative curvature \cite{nickel2017poincare, chamberlain2017neural}. This property has led to successful applications across various modalities with inherent hierarchies, including graphs \cite{liu2019hyperbolic}, text \cite{dhingra2018embedding, tifrea2019poincare}, and images \cite{atigh2022hyperbolic}. Recent works have further extended hyperbolic geometry to the multi-modal domain. For instance, \citet{desai2023hyperbolic} propose hyperbolic vision-language models where image embeddings are constrained to lie within a concept cone defined by the text embeddings. Similarly, \citet{pal2024compositional} extend this framework to model hierarchical relations across multiple levels, including cropped-text, cropped-image, original text, and original image representations. Building on these insights, we extend hyperbolic representation learning to the domain of brain decoding. Specifically, we leverage the geodesic property of hyperbolic space to unify and compress semantic and perceptual visual information, providing a new perspective on aligning neural signals with rich visual stimuli.

\begin{figure*}[t]
    \centering
    \includegraphics[width=0.95\linewidth]{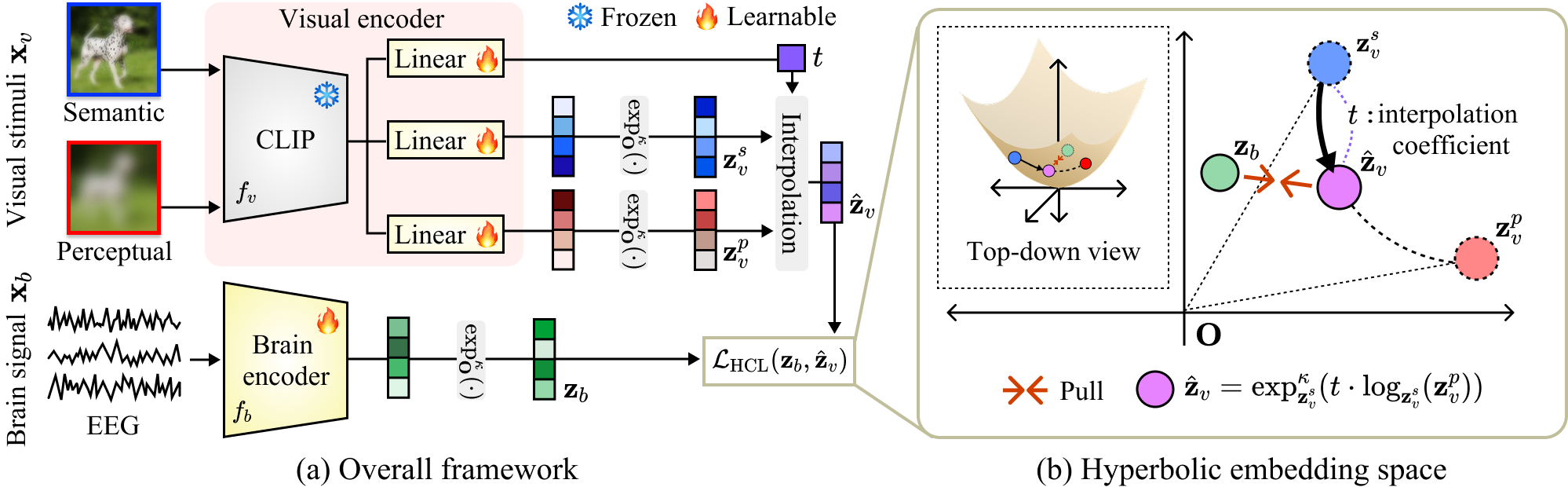}
    \caption{(a) The semantic image $\mathbf{x}_v^{s}$ and perceptual image $\mathbf{x}_v^{p}$ are encoded by CLIP and projected via a linear layer, and then lifted onto the hyperboloid via the exponential map. Using a learned weight $t$ derived from the semantic image features, the two image features are interpolated on the hyperbolic manifold. Similarly, EEG inputs are encoded and projected onto the same hyperbolic space. Contrastive learning is then performed on the hyperboloid to bring paired EEG-image representations closer. (b) A schematic view of the hyperbolic embedding space. The interpolated representation $\hat{\mathbf{z}}_v$ lies along the geodesic between the semantic feature $\mathbf{z}_v^{s}$ and the perceptual feature $\mathbf{z}_v^{p}$. Contrastive learning then pulls the EEG feature $\mathbf{z}_b$ toward the target $\hat{\mathbf{z}}_v$.
 }
    \label{framework}
\end{figure*}

\section{Preliminaries}\label{sec:pre}
We formulate our approach on hyperbolic space, a Riemannian manifold with constant negative curvature, where volume grows exponentially with radius. This property makes it well suited for representing hierarchical structures and addressing modality imbalance in multi-modal learning \citep{le2019inferring, peng2021hyperbolic}. Following prior works \citep{desai2023hyperbolic, pal2024compositional}, we adopt the Lorentz (hyperboloid) model due to its strong empirical performance in multi-modal tasks.

\subsubsection{Definition} The Lorentz model $\mathbb{L}^n$ of $n$-dimensional hyperbolic space with constant negative curvature is realized as the “upper sheet” of a two-sheeted hyperboloid in $(n+1)$-dimensional Minkowski space \cite{cannon1997hyperbolic}. Concretely, a point $\mathbf{p} \in \mathbb{R}^{n+1}$ is represented as $\mathbf{p}=(p_0,\tilde{\mathbf{p}})$, where $p_0 > 0$ denotes the time component and $\tilde{\mathbf{p}}\in\mathbb{R}^n$ are spatial coordinates. The Lorentz manifold is defined as: 
\begin{equation}
\label{lorentz}
\mathbb{L}^n = \left\{ \mathbf{p} \in \mathbb{R}^{n+1} : \langle \mathbf{p}, \mathbf{p} \rangle_{\mathbb{L}} = -\frac{1}{\kappa},\ p_0 > 0 \right\},
\end{equation}
where $ -\kappa \in \mathbb{R}$ is the curvature of the space. The Lorentzian inner product for two vectors  $\mathbf{p}, \mathbf{q} \in \mathbb{L}^n$ is defined as:
\begin{equation}
\langle \mathbf{p}, \mathbf{q} \rangle_{\mathbb{L}} = -p_0 q_0 + \langle \tilde{\mathbf{p}}, \tilde{\mathbf{q}} \rangle_{\mathbb{E}},  
\end{equation}
where $\langle \cdot, \cdot \rangle_{\mathbb{E}}$ denotes the standard Euclidean dot product.

\subsubsection{Geodesics} A geodesic is the shortest curve connecting two points on the manifold. In the Lorentz model, the geodesic distance between two points $\mathbf{p}, \mathbf{q} \in \mathbb{L}^n$ is defined as:
\begin{equation}
    d_{\mathbb{L}}(\mathbf{p}, \mathbf{q}) = \sqrt{1/\kappa} \cdot \cosh^{-1} \left( -\kappa \langle \mathbf{p}, \mathbf{q} \rangle_{\mathbb{L}} \right).
\end{equation}

\subsubsection{Exponential and Logarithmic Map}
The exponential map defines a smooth mapping from the tangent space onto the Lorentz manifold.
For a point $\mathbf{p} \in \mathbb{L}^n$, the tangent space is defined as:
\begin{equation} \label{tangent}
T_{\mathbf{p}} \mathbb{L}^n = \left\{ \mathbf{v} \in \mathbb{R}^{n+1} \mid \langle \mathbf{p}, \mathbf{v} \rangle_{\mathbb{L}} = 0 \right\}.
\end{equation}
Given a tangent vector $\mathbf{v} \in T_{\mathbf{p}} \mathbb{L}^n$, the exponential map traces the geodesic from $\mathbf{p}$ in the direction of $\mathbf{v}$ and is parameterized as $\gamma(t) = \exp_{\mathbf{p}}^{\kappa}(t\mathbf{v})$, where $t \in [0,1]$. It is explicitly defined as:
\begin{equation}\label{exponetial map}
\exp^\kappa_{\mathbf{p}}(t\mathbf{v}) = \cosh\left( t\sqrt{\kappa} \|\mathbf{v}\|_{\mathbb{L}} \right) \mathbf{p} + \frac{\sinh\left( t\sqrt{\kappa} \|\mathbf{v}\|_{\mathbb{L}} \right)}{\sqrt{\kappa} \|\mathbf{v}\|_{\mathbb{L}}} \mathbf{v},
\end{equation}
where $||\mathbf{v}||_{\mathbb{L}} = \langle \mathbf{v},\mathbf{v} \rangle_{\mathbb{L}}$. Conversely, a point $\mathbf{q} \in \mathbb{L}^n$ on the hyperboloid can be projected onto the tangent space via the logarithmic map $\log_{\mathbf{p}}^{\kappa}(\cdot): \mathbb{L}^n \to T_{\mathbf{p}}\mathbb{L}^n$, as follows:
\begin{equation}
\log_{\mathbf{p}}^{\kappa}(\mathbf{q}) = 
\frac{\cosh^{-1}(-\kappa \langle \mathbf{p}, \mathbf{q} \rangle_{\mathbb{L}})}
{\sqrt{(\kappa \langle \mathbf{p}, \mathbf{q} \rangle_{\mathbb{L}})^2 - 1}} 
\left( \mathbf{q} + \kappa \langle \mathbf{p}, \mathbf{q} \rangle_{\mathbb{L}} \mathbf{p} \right).
\end{equation}

In practice, the point $\mathbf{p}$ is commonly set to the time origin $ \mathbf{O} = (\sqrt{1/\kappa}, 0, \ldots, 0)^\top \in \mathbb{L}^{n}$. Under this setting, a vector $\mathbf{v} = [ 0, \mathbf{v}_{\text{enc}}] \in \mathbb{R}^{n+1}$ lies in the tangent space at $\mathbf{O}$ and can be mapped onto the hyperboloid via exponential map $\exp_{\mathbf{O}}^{\kappa}(\cdot)$, where $\mathbf{v}_{\text{enc}}$ denotes the encoder output.

\section{Method}
\subsection{Problem Formulation}
Given the brain signals space $\mathcal{X}_b$ and the visual stimuli space $\mathcal{X}_v$, the goal of visual brain decoding is to map brain signals $\mathbf{x}_b \in \mathcal{X}_b$ into a shared space $\mathcal{H}$ aligned with $\mathbf{x}_v \in \mathcal{X}_v$. To this end, we learn a brain encoder $f_b: \mathcal{X}_b \to \mathcal{H}$ and a visual encoder $Wf_v: \mathcal{X}_v \to \mathcal{H}$, where $f_v$ is a frozen backbone of pre-trained VLM and $W$ is a linear layer. Specifically, we use $n$-dimensional Lorentz space $\mathbb{L}^{n}$ as a semantically aligned space $\mathcal{H}$. 
Embeddings in $\mathbb{L}^n$ are obtained via the exponential map $\exp_{\mathbf{O}}^{\kappa}(\cdot)$ at the time origin $\mathbf{O}$. The visual embedding is defined as $\mathbf{z}_v = \exp_{\mathbf{O}}^{\kappa}(\alpha_v \cdot W f_v(\mathbf{x}_v))$ and the brain embedding $\mathbf{z}_b = \exp_{\mathbf{O}}^{\kappa}(\alpha_b\cdot f_b(\mathbf{x}_b))$, where $\alpha_v$ and $\alpha_b$ are learnable projection scalars that reduce the norm of the embeddings to keep them near the origin $\mathbf{O}$. An overview of our overall framework is illustrated in Fig~\ref{framework}(a).

\subsection{Hyperbolic Brain-Vision Contrastive Learning}
To align neural embeddings with corresponding visual embedding, we utilize contrastive learning in hyperbolic space. Unlike Euclidean space, hyperbolic space provides a natural embedding space for aligning modalities with asymmetric information capacity and has shown empirical success in multi-modal representation learning \citep{desai2023hyperbolic, pal2024compositional, mandica2025hyperbolic}.

Given a batch of EEG-image pairs $\{(\mathbf{z}_{b,i}, \mathbf{z}_{v,i})\}_{i=1}^{B}$, hyperbolic contrastive learning is formulated as:
\begin{equation}\label{contrastive_learning}
        \mathcal{L}(
        {\mathbf{z}}_v, \mathbf{z}_b) = - \sum_{i \in B} \log \frac{
\exp\left(d_{\mathbb{L}}(\mathbf{z}_{v,i}, \mathbf{z}_{b,i})/ \tau\right)
}{
\sum\limits_{k = 1,\, k \ne i}^{B} 
\exp\left(d_{\mathbb{L}}(\mathbf{z}_{v,i}, \mathbf{z}_{b,k}) / \tau\right)
},
\end{equation}
where $B$ denote the batch, $d_\mathbb{L}$ denote the negative Lorentz distance, $\tau$ is temperature parameter.

\subsection{Hyperbolic Feature Interpolation}
With hyperbolic space established, we aim to learn visual representations aligned with brain signals by capturing their inherent properties. In particular, we fuse and compress semantic and perceptual visual features. Our approach is motivated by two key observations: (1) semantic and perceptual visual features are often entangled in neural activity, (2) and brain signals inherently contain less information than natural images. To address the first property, we describe how semantic and perceptual features are extracted from an image, then present a hyperbolic interpolation method. Finally, we show that this interpolation naturally leads to information compression, thereby addressing the second observation.

\subsubsection{Extracting Semantic and Perceptual Features} 
\begin{figure}[t]
    \centering
    \includegraphics[width=1\linewidth]{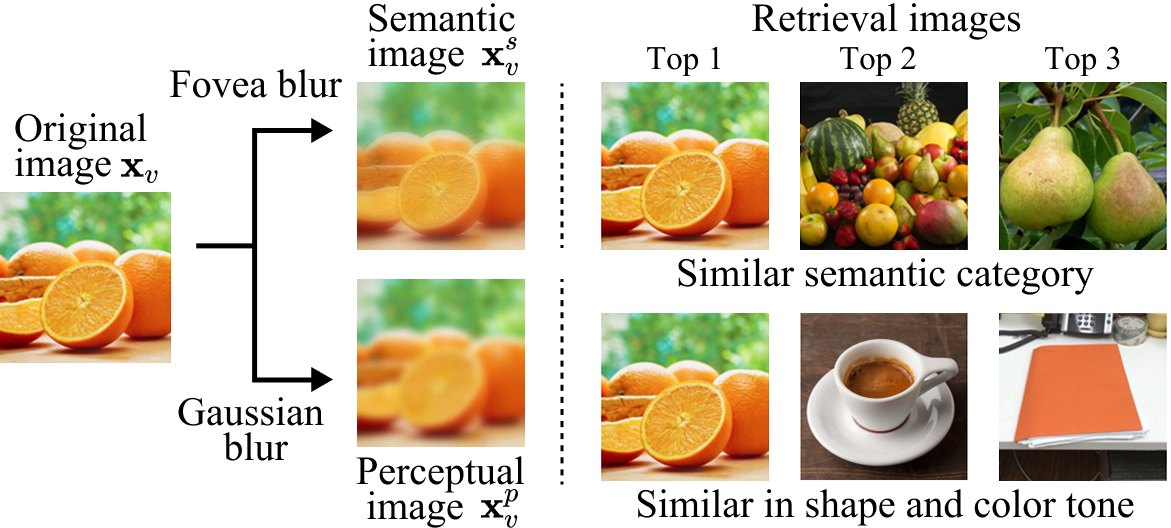}
    \caption{Examples of image augmentations and retrieval results. 
    The semantic image $\mathbf{x}_v^{s}$ and perceptual image $\mathbf{x}_v^{p}$ are generated via fovea blur and Gaussian blur, respectively. Retrieval results using CLIP embedding show that semantic queries return category-relevant matches (e.g., fruits),  while perceptual queries retrieve images with similar low-level attributes such as color and shape.
 }
    \label{aug_example}
\end{figure}

We first apply image-level augmentations to obtain the semantic and perceptual visual inputs. The semantic image $\mathbf{x}_v^{s}$ is generated via fovea blur, simulating peripheral vision to preserve semantics and enhance alignment with brain signals \citep{wu2025bridging}. The perceptual image $\mathbf{x}_v^{p}$ is obtained by applying Gaussian blurring to suppress high-frequency components and retain coarse structure. This augmentation amplifies perceptual attributes in the CLIP embeddings. These augmentations and their effects are presented in Fig.~\ref{aug_example}.

These inputs are then projected into hyperbolic space as semantic visual features $\mathbf{z}_v^{s} = \exp_{\mathbf{O}}^{\kappa}(\alpha_v\cdot W_sf_v(\mathbf{x}_v^{s}))$ and perceptual visual features $\mathbf{z}_v^{p} = \exp_{\mathbf{O}}^{\kappa}(\alpha_v \cdot W_pf_v(\mathbf{x}_v^{p}))$, respectively. Here, $W_s, W_p \in \mathbb{R}^{d\times d}$ are learnable matrices and $d$ denotes the CLIP embedding dimension.


\subsubsection{Hyperbolic Interpolation} To perform interpolation of these features, we approximate the geodesic in the Lorentz model using the exponential map. The perceptual feature $\mathbf{z}_v^{p}$ is projected onto the tangent space at $\mathbf{z}_v^{s}$ using the logarithmic map, i.e., $\log^{\kappa}_{\mathbf{z}_v^{s}}(\mathbf{z}_v^p) \in T_{\mathbf{z}_v^{s}}\mathbb{L}^{n}$. This tangent vector is scaled and mapped back to the hyperbolic space via the exponential map. The resulting interpolated visual representation $\hat{\mathbf{z}}_v$ follows the geodesic from $\mathbf{z}_v^{s}$ to $\mathbf{z}_v^{p}$:
\begin{equation}
    \hat{\mathbf{z}}_v = \gamma_{\mathbf{z}_v^{s} \to \mathbf{z}_v^{p}}(t) = \exp^{\kappa}_{\mathbf{z}_v^{s}}\left(t \cdot \log^{\kappa}_{\mathbf{z}_v^{s}}(\mathbf{z}_v^{p})\right),
\label{geodesic}
\end{equation}
where $t \in [0, 1]$ is the interpolation coefficient.

We compute the interpolation coefficient $t$ dynamically to reflect image-specific variation in the relative importance of semantic and perceptual features:
\begin{equation}
t = \sigma(W_t f_v(\mathbf{x}_v^{s})),
\end{equation}
where $ W_t \in \mathbb{R}^{1 \times d}$ is a learnable matrix and $\sigma$ denotes the sigmoid function.

\subsubsection{Compression Effects} The proposed interpolation mechanism concurrently compresses and fuses semantic and perceptual features. To analyze how this compression arises, we revisit the geodesic formulation in the Lorentz model. We reformulate the geodesic in Eq. (\ref{geodesic}) as follows: 
\begin{equation}
    \gamma_{\mathbf{p} \to \mathbf{q}}(t) = \frac{\sinh\left((1 - t)\beta\right)}{\sinh(\beta)} \mathbf{p} + \frac{\sinh\left(t \beta\right)}{\sinh(\beta)} \mathbf{q},
\end{equation}
where $\beta = \sqrt\kappa \cdot d_{\mathbb{L}}(\mathbf{p}, \mathbf{q})$. A detailed derivation is provided in the appendix.

Unlike linear interpolation $(1 - t)\mathbf{p} + t\mathbf{q}$ in Euclidean space, the hyperbolic interpolation weights $\frac{\sinh((1 - t)\beta)}{\sinh(\beta)}$ and $\frac{\sinh(t\beta)}{\sinh(\beta)}$ are strictly smaller than $(1 - t)$ and $t$, respectively. This causes the interpolated points to lie closer to the origin.

To understand how this contraction relates to the geometry of hyperbolic space, we revisit the hyperboloid constraint in Eq.~(\ref{lorentz}), which gives:
\begin{equation}
    p_0 = \sqrt{1/\kappa + \|\tilde{\mathbf{p}}\|^2},
\label{eq:p_0}
\end{equation}
A smaller $p_0$ constrains $\|\tilde{\mathbf{p}}\|$ more tightly, thus reducing the expressive capacity of embedding. This is consistent with prior work \citep{ganea2018hyperbolic, khrulkov2020hyperbolic}, suggesting that points near the origin represent more abstract concepts.

\subsubsection{Final Objective Function}
Finally, we train our model by aligning the interpolated visual representations with the brain embeddings in hyperbolic space.
The final hyperbolic contrastive loss is defined as:
\begin{equation}
\mathcal{L}_{\mathrm{HCL}} = \mathcal{L}(\hat{\mathbf{z}}_v, \mathbf{z}_b) + \mathcal{L}(\mathbf{z}_b, \hat{\mathbf{z}}_v).
\end{equation}

\begin{table*}[t]
\footnotesize 
\setlength{\tabcolsep}{2.7pt}
\centering
\begin{tabular}{lcccccccccccccccccccccccc}
\toprule
Method & \multicolumn{2}{c}{Subject 1} & \multicolumn{2}{c}{Subject 2} & \multicolumn{2}{c}{Subject 3} & \multicolumn{2}{c}{Subject 4} & \multicolumn{2}{c}{Subject 5} & \multicolumn{2}{c}{Subject 6} & \multicolumn{2}{c}{Subject 7} & \multicolumn{2}{c}{Subject 8} & \multicolumn{2}{c}{Subject 9} & \multicolumn{2}{c}{Subject 10} & \multicolumn{2}{c}{Average} \\
& T-1 & T-5 & T-1 & T-5 & T-1 & T-5 & T-1 & T-5 & T-1 & T-5 & T-1 & T-5 & T-1 & T-5 & T-1 & T-5 & T-1 & T-5 & T-1 & T-5 & T-1 & T-5
\\
\midrule
\midrule
\multicolumn{23}{c}{\textbf{Intra-subject:} train and test on one subject} \\
\midrule
BraVL   & 6.1 & 17.9 & 4.9 & 14.9 & 5.6 & 17.4 & 5.0 & 15.1 & 4.0 & 13.4 & 6.0 & 18.2 & 6.5 & 20.4 & 8.8 & 23.7 & 4.3 & 14.0 & 7.0 & 19.7 & 5.8 & 17.5 \\
NICE    & 13.2 & 39.5 & 13.5 & 40.3 & 14.5 & 42.7 & 20.6 & 52.7 & 10.1 & 31.5 & 16.5 & 44.0 & 17.0 & 42.1 & 22.9 & 56.1 & 15.4 & 41.6 & 17.4 & 45.8 & 16.1 & 43.6 \\
ATM-S   & 25.6 & 60.4 & 22.0 & 54.5 & 25.0 & 62.4 & 31.4 & 60.9 & 12.9 & 43.0 & 21.3 & 51.1 & 30.5 & 61.5 & 38.8 & 72.0 & 30.4 & 51.5 & 29.1 & 63.5 & 28.5 & 60.4 \\
Cog-cap & 31.4 & 79.7 & 31.4 & 77.8 & 38.2 & 85.7 & 40.4 & 85.8 & 24.4 & 66.3 & 34.8 & 78.8 & 34.7 & 81.0 & 48.1 & 88.6 & 37.4 & 79.4 & 35.6 & 79.3 & 35.6 & 80.2 \\
UBP & \underline{41.2} & \underline{70.5} & \underline{51.2} & \underline{80.9} & \underline{51.2} & \underline{82.0} & \underline{51.1} & \underline{76.9} & \underline{42.2} & \underline{72.8} & \underline{57.5} & \underline{83.5} & \underline{49.0} & \underline{79.9} & \underline{58.6} & \underline{85.8} & \underline{45.1} & \underline{76.2} & \underline{61.5} & \underline{88.2} & \underline{50.9} & \underline{79.7} \\

\midrule
HyFI & \textbf{60.6} & \textbf{85.3} & \textbf{65.9} & \textbf{94.0} & \textbf{69.5} & \textbf{93.9} & \textbf{66.5} & \textbf{89.8} & \textbf{55.0} & \textbf{86.0} & \textbf{74.4} & \textbf{95.0} & \textbf{68.4} & \textbf{91.3} & \textbf{78.9} & \textbf{96.9} & \textbf{66.0} & \textbf{90.6} & \textbf{77.0} & \textbf{96.4} & \textbf{68.2} & \textbf{91.9} \\

\midrule
\midrule
\multicolumn{23}{c}{\textbf{Inter-subject:} leave one subject out for test} \\
\midrule
BraVL & 2.3 & 8.0 & 1.5 & 6.3 & 1.9 & 6.7 & 2.1 & 8.1 & 2.2 & 7.6 & 1.6 & 6.4 & 2.3 & 8.5 & 1.8 & 7.0 & 1.4 & 5.9 & 1.7 & 6.7 & 1.5 & 5.6 \\
NICE & 7.6 & 22.8 & 5.9 & 20.5 & 6.0 & 22.3 & 6.3 & 20.7 & 4.4 & 18.3 & 5.6 & 22.2 & 5.6 & 19.7 & 6.3 & 22.0 & 5.7 & 17.6 & 8.4 & 28.3 & 6.2 & 21.4 \\
NICE-G & 5.9 & 21.4 & 6.4 & 22.7 & 5.5 & 20.1 & 6.1 & 21.0 & 4.7 & 19.5 & 6.2 & 22.5 & 5.9 & 19.1 & 7.3 & 25.3 & 6.2 & 18.3 & 6.2 & 26.3 & 5.9 & 21.6 \\
ATM-S & 10.5 & 26.8 & 7.1 & 24.8 & \textbf{11.9} & \textbf{33.8} & \underline{14.7} & \underline{39.4} & 7.0 & 23.9 & 11.1 & \textbf{35.8} & \textbf{16.1} & \textbf{43.5} & \textbf{15.0} & \textbf{40.3} & 4.9 & 22.7 & \underline{20.5} & \underline{46.5} & 11.8 & \underline{33.7} \\
UBP & \underline{11.5} & \underline{29.7} & \underline{15.5} & \underline{40.0} & \underline{9.8} & \underline{27.0} & 13.0 & 32.3 & \underline{8.8} & \textbf{33.8} & \underline{11.7} & 31.0 & 10.2 & 23.8 & 12.2 & \underline{32.2} & \textbf{15.5} & \textbf{40.5} & 16.0 & 43.5 & 12.4 & 33.4 \\
\midrule
HyFI & \textbf{16.2} & \textbf{35.8} & \textbf{20.0} & \textbf{47.7} & 7.5 & 26.7 & \textbf{18.8} & \textbf{41.3} & \textbf{9.7} & \underline{27.5} & \textbf{15.5} & \underline{33.8} & \underline{11.0} & \underline{34.5} & \underline{13.2} & 30.3 & \underline{13.3} & \underline{38.8} & \textbf{25.3} & \textbf{55.7} & \textbf{15.1} & \textbf{37.2} \\
\midrule
\bottomrule
\end{tabular}
\caption{Top-1 (T-1) and top-5 (T-5) accuracy ($\%$) results in 200-way zero-shot brain-to-image retrieval on THINGS-EEG, reported for intra- and inter-subject settings. \textbf{Bold} and \underline{underline} indicate the best and second-best results, respectively.}
\label{tab:things_eeg_main}
\end{table*}

\begin{table}[t] 
\footnotesize 
\setlength{\tabcolsep}{2.5
pt}
\centering
\begin{tabular}{lcccccccccc}
\toprule
Method & \multicolumn{2}{c}{Subject 1} & \multicolumn{2}{c}{Subject 2} & \multicolumn{2}{c}{Subject 3} & \multicolumn{2}{c}{Subject 4} & \multicolumn{2}{c}{Average}\\
 & T-1 & T-5 & T-1 & T-5 & T-1 & T-5 & T-1 & T-5 & T-1 & T-5 \\
 \hline
\midrule
\multicolumn{11}{c}{\textbf{Intra-subject:} train and test on one subject}
\\
\midrule
NICE & 8.7  & 30.5 & 21.8 & 56.6 & 16.5 & 49.7 & 10.3 & 32.3 & 14.3 & 42.3 \\
UBP & \underline{15.0} & \underline{38.0} & \underline{46.0} & \underline{80.5} & \underline{27.3} & \underline{59.0} & \underline{18.5} & \underline{43.5} & \underline{26.7} & \underline{55.2} \\
\midrule

\textbf{HyFI} & \textbf{17.6} & \textbf{40.1} & \textbf{63.8} & \textbf{91.1} & \textbf{38.0} & \textbf{76.9} & \textbf{23.7} & \textbf{50.2} & \textbf{35.8} & \textbf{64.6} \\

\midrule
\midrule
\multicolumn{11}{c}{\textbf{Inter-subject:} leave one subject out for test} \\
\midrule
UBP & 2.0  & 5.7  & 1.5  & 17.2 & 2.7  & 10.5 & \textbf{2.5}  & \textbf{8.0}  & 2.2  & 10.4 \\
\midrule
\textbf{HyFI} & \textbf{2.7} & \textbf{6.0} & \textbf{4.3} & \textbf{17.0} & \textbf{3.5} & \textbf{12.5} & \underline{1.0} & \underline{7.6} & \textbf{3.2} & \textbf{11.5} \\
\midrule
\bottomrule
\end{tabular}
\caption{Top-1 and top-5 accuracy ($\%$) results in 200-way zero-shot brain-to-image retrieval on THINGS-MEG.}
\label{tab:things_meg_main}

\end{table}

\section{Experiments and Results}

\subsection{Datasets}
\subsubsection{THINGS-EEG} We used the THINGS-EEG dataset \cite{gifford2022large}, a large-scale EEG benchmark collected under the Rapid Serial Visual Presentation (RSVP) paradigm, which provides image-EEG paired data from 10 subjects viewing rapid sequences of images. The training set consists of 1,654 object concepts, each represented by 10 distinct images, with each image repeated 4 times per subject. The test set includes 200 concepts with one image per concept and each image repeated 80 times per subject. Following prior preprocessing protocols \citep{songdecoding,wu2025bridging}, we averaged trials, downsampled EEG signals to 250 Hz, and selected 17 channels over occipital and parietal regions.

\subsubsection{THINGS-MEG} We additionally utilized the THINGS-MEG dataset~\cite{hebart2023things}, which contains recordings from four participants using 271 MEG channels. The training set includes 1,854 object concepts, each presented once with 12 different images, while the test set consists of 200 concepts, each associated with a single image repeated 12 times. To ensure fair comparison, we follow the same preprocessing pipeline as described in~\cite{songdecoding, wu2025bridging}; more details are provided in the appendix.


\subsection{Implementation Details}
We used AdamW~\cite{loshchilov2017decoupled} with learning rate $3 \times 10^{-4}$, weight decay $1 \times 10^{-4}$, and batch size 1024. We trained the model for 50 epochs. The curvature parameter $\kappa$ was set to 1 at initialization and optimized during training. To control feature norms, we initialize the scaling factors $\alpha_v$ and $\alpha_b$ to $\sqrt{1/d}$, following ~\cite{desai2023hyperbolic}.
We empirically observed that fixing $\alpha_v = 1$ led to better training stability for ResNet-based and hyperbolic vision encoder, due to the already exhibit low norm of their features. All experiments  were conducted on a GPU, GTX 1080 Ti (12GB). We applied fovea blur and Gaussian blur. Augmentation details are in the appendix. 

\subsubsection{Vision Encoders} 
We evaluated a range of visual backbones, including RN50, RN101, ViT-B/16, ViT-B/32, ViT-L/14, and ViT-H/14. In addition, we also used two recent hyperbolic vision-language models, MERU~\cite{desai2023hyperbolic} and HyCoCLIP~\cite{pal2024compositional}, which are pretrained to capture hierarchical relationships between images and text. We adopted RN50 as the default vision encoder.  

\subsubsection{Brain Encoders}
We applied our method using a variety of EEG encoders to demonstrate its generality across different architectures. Specifically, we adopted models that are either widely used in the literature or recently proposed in neural encoding studies, including ShallowNet~\cite{schirrmeister2017deep}, EEGNet~\cite{lawhern2018eegnet}, and TSConv~\cite{li2024visual}, as well as EEGProject~\cite{wu2025bridging}. We used EEGProject as default EEG encoder.

\subsection{Results}

We compare our method with several recent neural decoding approaches for brain-to-image retrieval, including BraVL~\cite{du2023decoding}, NICE~\cite{songdecoding}, ATM~\cite{li2024visual}, CogCap~\cite{zhang2025cognitioncapturer}, and UBP~\cite{wu2025bridging}. Detailed baseline descriptions are in the appendix.

To evaluate decoding performance, we perform 200-way zero-shot retrieval on THINGS-EEG and THINGS-MEG, following prior work~\cite{du2023decoding, wu2025bridging}. This evaluates alignment quality and generalization to novel concepts. The results in Tables~\ref{tab:things_eeg_main} and~\ref{tab:things_meg_main}.
We averaged over 5 runs, and all improvements are statistically significant ($p < 0.01$). 
On the THINGS-EEG dataset, our method achieves a top-1 accuracy of 68.2$\%$ and top-5 accuracy of 91.9$\%$, outperforming the previous SOTA (UBP) by +17.3$\%$ and +12.2$\%$, respectively. On the THINGS-MEG dataset, HyFI achieves a top-1 accuracy of 35.8$\%$ and top-5 accuracy of 64.6$\%$, improving upon UBP by +9.1$\%$ and +9.4$\%$, respectively.

\begin{figure}[h]
    \centering
    \includegraphics[width=1\linewidth]{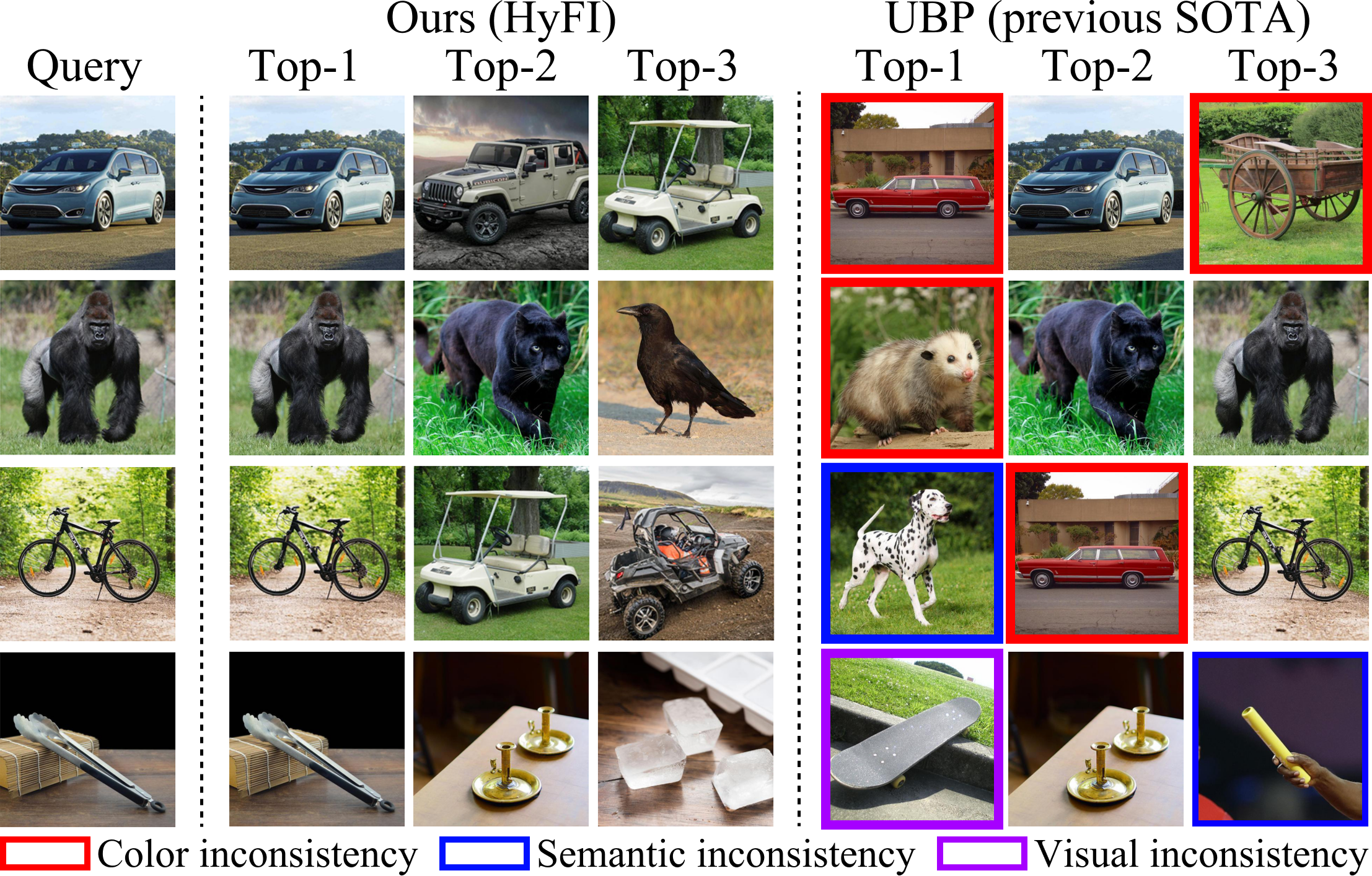}
\caption{Qualitative comparison of image retrieval results. Our method retrieves semantically and perceptually coherent images, while the previous method often suffers from color or semantic inconsistencies.}
\label{compare}
\end{figure}

Furthermore, we conduct a qualitative comparison of retrieval results with UBP, as shown in Fig.~\ref{compare}. For fair comparison, we used subject 4 with near-average performance. As illustrated in the figure, our method preserves both semantic and perceptual coherence, whereas previous approaches fail to maintain this consistency. This indicates that our method effectively fuses perceptual and semantic features, thereby maintaining coherence in both aspects. More examples of retrieved image are provided in the appendix.

\subsection{Ablation Study}
\subsubsection{Effect of Hyperbolic Geometry and Feature Interpolation}
To evaluate the contributions of hyperbolic space and feature interpolation, we conduct an ablation study and results in Table~\ref{tab:abl}. First, we observe that aligning brain and image features is more effective in hyperbolic space than in CLIP space (Euclidean space). This supports the suitability of hyperbolic space for modeling the information imbalance between brain and visual modalities. Next, we compare interpolating features in the original CLIP space and in hyperbolic space. Even in the CLIP space, fusing low-level visual features improves performance, highlighting the benefit of integrating perceptual information. However, performing interpolation in hyperbolic space leads to the best results. This finding indicates that hyperbolic interpolation effectively reduces redundancy and representational complexity, while facilitating feature fusion.

\begin{table}[t]
\centering
\small
\setlength{\tabcolsep}{2.5pt}
\begin{tabular}{cccccc}
\toprule
\multirow{2}{*}{Interpolation} & \multirow{2}{*}{Hyperbolic} & \multicolumn{2}{c}{THINGS-EEG} & \multicolumn{2}{c}{THINGS-MEG} \\
 & & T-1 & T-5 & T-1 & T-5 \\
\midrule
\midrule
-- & -- & 49.4 & 81.0 & 23.1 & 47.8 \\
-- & \checkmark & 54.3 & 82.5 & 28.8 & 55.1 \\
\checkmark & -- & 59.7 & 86.8 & 25.2 & 51.8 \\
\checkmark & \checkmark & \textbf{68.2} & \textbf{91.9} & \textbf{35.8} & \textbf{64.6} \\
\midrule
\bottomrule
\end{tabular}
\caption{Ablation study: effect of interpolation and hyperbolic space. The row with only interpolation checked refer to interpolation in CLIP space.}
\label{tab:abl}
\end{table}

\begin{table}[t]
\centering
\setlength{\tabcolsep}{3pt}

\footnotesize
\setlength{\tabcolsep}{1.0pt}
\begin{tabular}{c c c c c c c}
\toprule
\multirow{2}{*}{Method} & \multirow{2}{*}{Architecture} & \multirow{2}{*}{Model} 
& \multicolumn{2}{c}{THINGS-EEG}
& \multicolumn{2}{c}{THINGS-MEG} \\
\cmidrule(lr){4-5} \cmidrule(lr){6-7}
& & & w/o HyFI & Ours & w/o HyFI & Ours \\
\midrule
\midrule

\multirow{6}{*}{CLIP} & \multirow{2}{*}{CNN} & RN50        & 49.4   & \textbf{68.2} & 23.1 & \textbf{35.8} \\
                      &                      & RN101       & 44.4   & \textbf{62.1} & 21.0 & \textbf{34.8} \\
\cmidrule(lr){2-7}
                      & \multirow{4}{*}{ViT} & ViT-B/16    & 36.0  & \textbf{41.6} & 18.6 & \textbf{23.3} \\
                      &                      & ViT-B/32    & 42.5   & \textbf{46.9} & 20.0 & \textbf{25.1} \\
                      &                      & ViT-L/14    & 30.1   & \textbf{34.1} & 13.0 & \textbf{21.0} \\
                      &                      & ViT-H/14    & 42.5   & \textbf{43.8} & 18.7 & \textbf{27.9} \\
\midrule
\multirow{3}{*}{MERU} & \multirow{3}{*}{ViT} &  ViT-S/16     & 43.7   & \textbf{52.4} & 21.6 & \textbf{31.1} \\
                            &                      & ViT-B/16    & 31.7   & \textbf{48.8} & 15.7 & \textbf{24.6} \\
                            &                      & ViT-L/16    & 23.1   & \textbf{30.9} & 11.0 & \textbf{19.6} \\
\midrule
\multirow{2}{*}{HyCoCLIP} 
                      & \multirow{2}{*}{ViT} & ViT-S/16 & 44.3   & \textbf{51.6} & 21.2 & \textbf{27.9} \\
                      &                      & ViT-B/16 & 35.6   & \textbf{45.6} & 18.1 & \textbf{27.0} \\
\midrule
\bottomrule
\end{tabular}
\caption{ Top-1 image retrieval accuracy ($\%$)  across visual encoders. The “w/o HyFI” represents alignment in CLIP space; MERU and HyCoCLIP use hyperbolic alignment.}
\label{tab:vision}
\end{table}

\begin{table}[t]
\centering
\footnotesize
\begin{tabular}{lcccc}
\toprule
\multirow{2}{*}{Brain Encoder} 
& \multicolumn{2}{c}{THINGS-EEG}
& \multicolumn{2}{c}{THINGS-MEG} \\
\cmidrule(lr){2-3} \cmidrule(lr){4-5}
& w/o HyFI & Ours 
& w/o HyFI & Ours \\
\midrule
\midrule
ShallowNet & 36.6 & \textbf{50.5} & 15.1 & \textbf{23.2} \\
EEGNet     & 36.2 & \textbf{51.6} & 19.5 & \textbf{26.4} \\
TSConv     & 41.1 & \textbf{57.1} & 21.9 & \textbf{30.5} \\
EEGProject & 49.4 & \textbf{68.2} & 23.1 & \textbf{35.8} \\
\midrule
\bottomrule
\end{tabular}
\caption{Top-1 image retrieval accuracy (\%) using different brain encoders and CLIP-RN50.}
\label{tab:brain encoder}
\end{table}

\subsection{Effect of Vision and Brain Encoders}
We investigate the impact of different vision and brain encoders on brain-to-image retrieval performance, the results summarized in Table~\ref{tab:vision} and \ref{tab:brain encoder}. For this analysis, we use EEGProject as the brain encoder and compare various pretrained vision encoders. We observe that CNN-based backbones yield stronger alignment with EEG signals compared to transformer-based models. Interestingly, lightweight architectures outperform deeper models. These results suggest that compact visual representations may be more compatible with neural signals.
Importantly, our method consistently outperforms the baseline across all vision architectures. We also evaluate the effect of different brain encoders while fixing the vision encoder to CLIP-RN50. We observe that our method consistently improves performance across all brain encoder architectures. This highlights the general applicability and robustness of our approach, regardless of the specific choice of brain encoder.

\subsection{Analysis}
\subsubsection{Feature Visualization} 
\begin{figure}[h]
    \centering
    \includegraphics[width=1\linewidth]{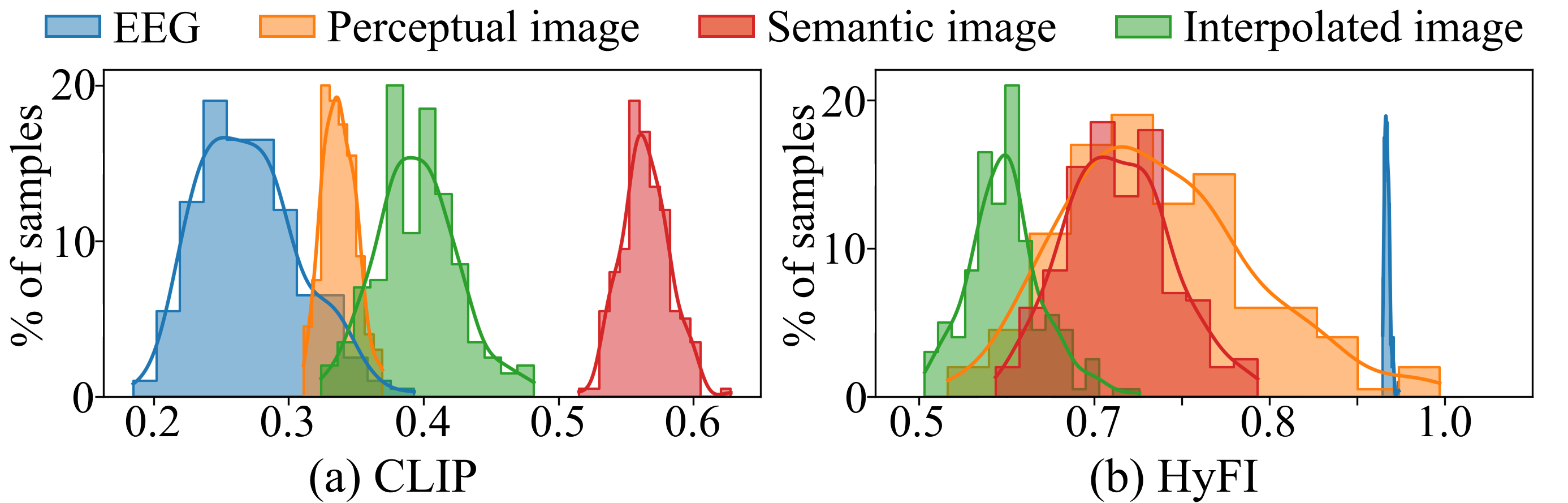}
\caption{Distributions of embedding distances from the root in (a) CLIP space and (b) hyperbolic space. Interpolated image embeddings lie closer to the root in hyperbolic space, unlike in CLIP space. }
\label{norm}
\end{figure}

To investigate the effect of feature interpolation in hyperbolic space, we visualize the distribution of distances from the root, as shown in Fig.~\ref{norm}.  Following~\cite{desai2023hyperbolic}, we define the root as the mean of all embeddings in CLIP space and the time origin \textbf{O} in hyperbolic space. As shown in Fig.~\ref{norm}, (a) in the CLIP space, the interpolated image embeddings lie between the semantic and perceptual features, whereas (b) in the hyperbolic space, the interpolated embeddings are located closer to the time origin \textbf{O}. This behavior reflects the nature of hyperbolic interpolation, which bends toward the origin and results in a tighter bound on the spatial components, as described in Eq.~(\ref{eq:p_0}). It suggests that effectively compressing and integrating semantic and perceptual features leads to better alignment with brain signals. Notably, the EEG embeddings lie farther from the origin, as their high variability requires regions with fewer constraints (i.e., away from the time origin).

\subsubsection{Analysis of interpolation coefficient} 
\begin{figure}[h]
    \centering
    \includegraphics[width=0.96\linewidth]{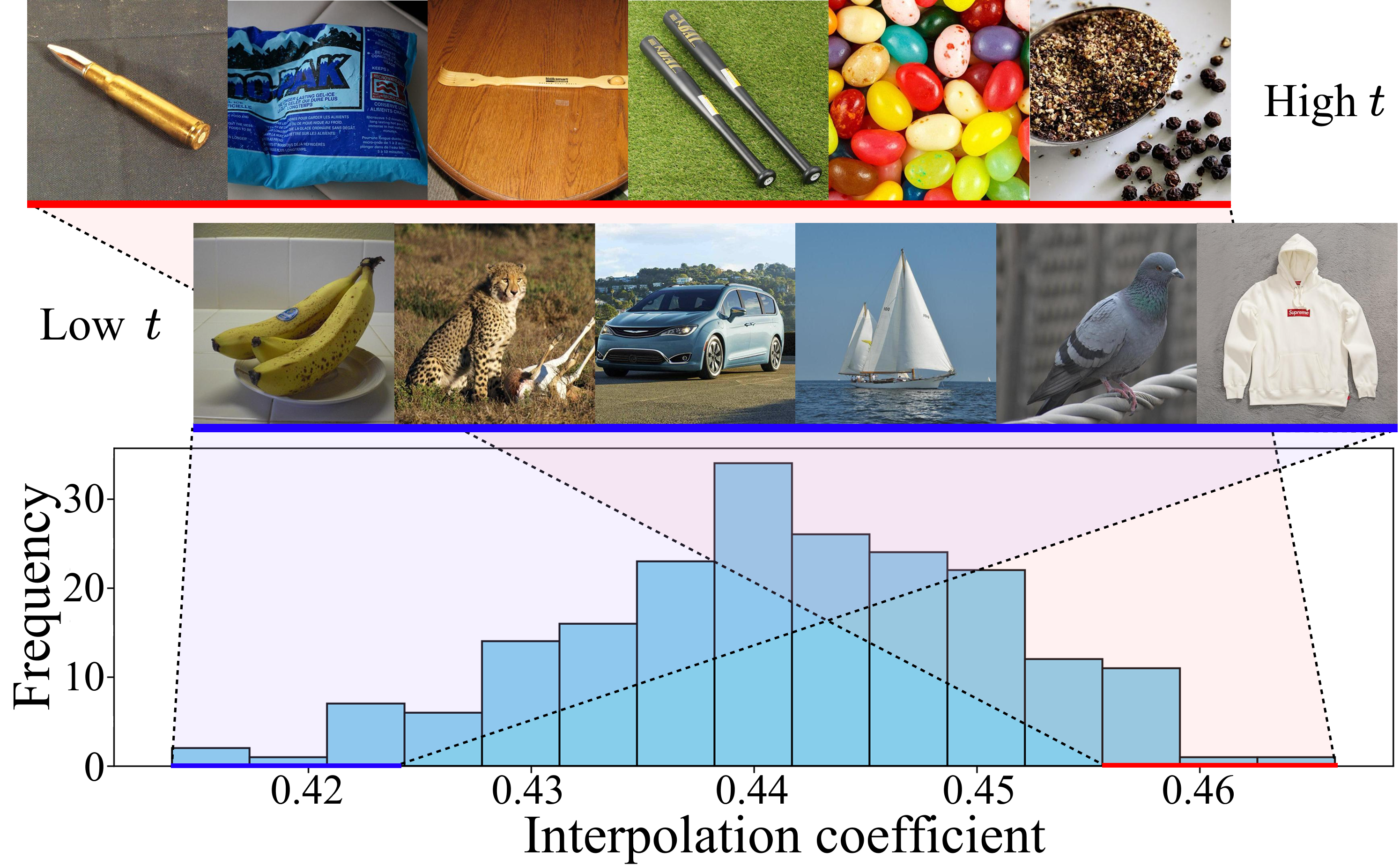}
    \caption{Distribution of the interpolation coefficient $t$ and example images with low and high $t$ values.
 }
\label{coefficient}
\end{figure}

To understand how the model adaptively integrates semantic and perceptual information, we analyze the learned interpolation coefficient $t \in [0,1]$. The distribution of coefficients is shown in Fig.~\ref{coefficient}. We observe that $t$ is predominantly distributed below 0.5, indicating that the model tends to focus more on semantic features during interpolation. To further interpret this behavior, we also examine images corresponding to both low and high $t$ values in the test set. Images with lower $t$ values typically contain objects that are iconic examples of their higher-level categories (e.g., banana–fruit, cheetah–mammal, and van-car). In contrast, images with higher $t$ values tend to exhibit salient low-level visual attributes such as orientation and color.

\begin{figure}[h]
    \centering
    \includegraphics[width=0.95\linewidth]{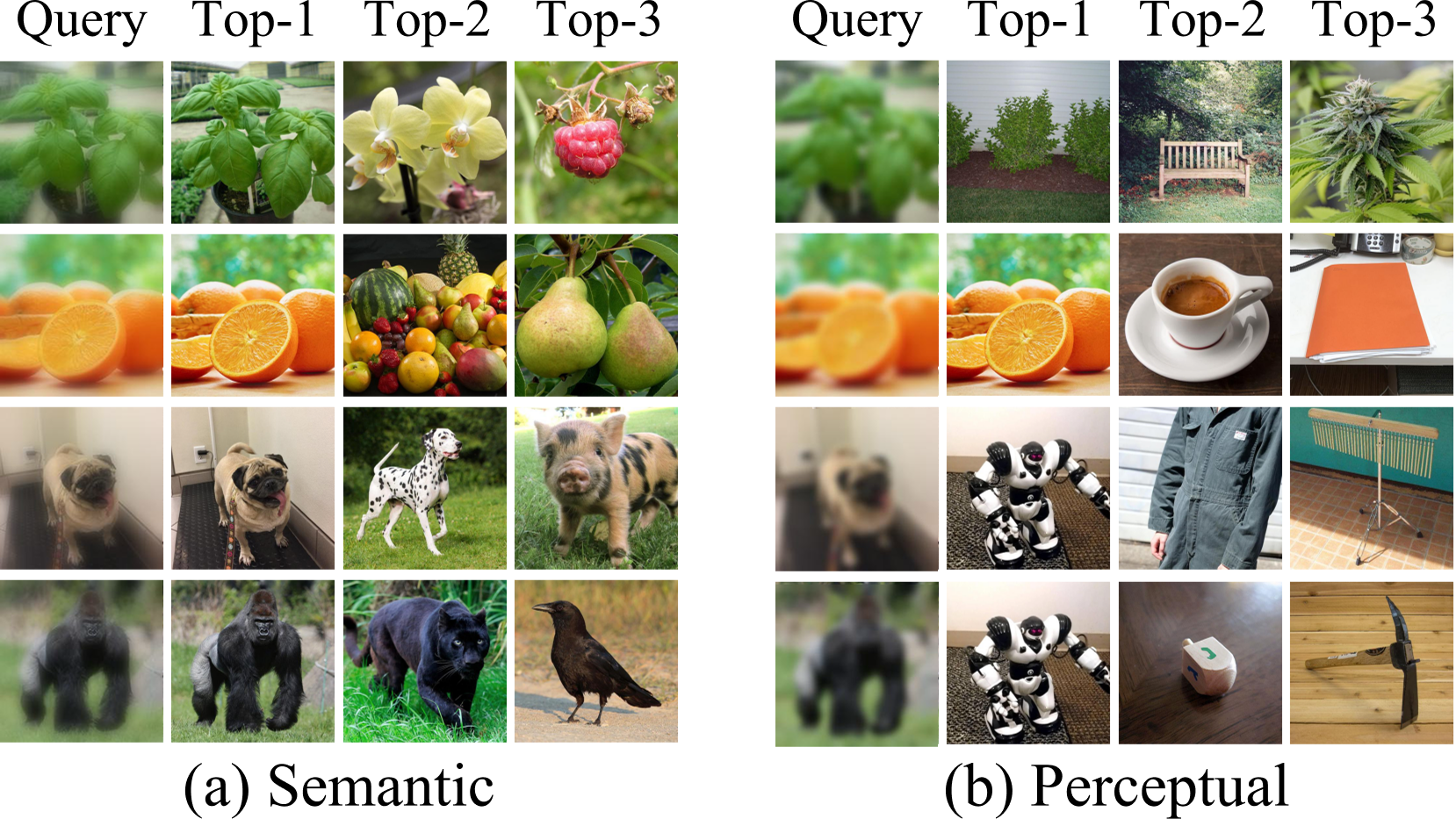}
    \caption{Comparison of Top-3 retrieval results using (a) semantic and (b) perceptual images. Semantic queries tend to retrieve conceptually similar images (e.g., plants, fruits, animals). In contrast, perceptual queries retrieve images with shared low-level visual feature such as color and orientation.
 }
\label{Aug_effect}
\end{figure}

\textbf{Analyzing the effect of image augmentation}
To investigate the effect of image augmentations on a pre-trained VLM, we conduct image-to-image retrieval using perceptual and semantic images as queries in CLIP-RN50. The results are shown in Fig.~\ref{Aug_effect}. When using the semantic images as queries, the retrieved results tend to be semantically consistent. In contrast, perceptual images bias the retrieval toward low-level visual attributes such as color and object orientation. Additional results with other VLMs are in the appendix.

\section{Conclusion}
We propose Hyperbolic Feature Interpolation (HyFI), a framework that interpolates semantic and perceptual visual features in hyperbolic space for improved alignment with brain signals. Leveraging the geodesic curvature toward the origin, HyFI enables effective fusion and compression of visual representations. This leads to better alignment with neural activity by reflecting both limited brain information and feature entanglement. Experiments on THINGS-EEG and THINGS-MEG show that HyFI achieves SOTA performance on zero-shot brain-to-image retrieval task.

\section*{Acknowledgments}
This work was supported by the Institute of Information \& Communications Technology Planning \& Evaluation (IITP) grant funded by the Korea government (MSIT) under two projects: (No. RS-2019-II190079, Artificial Intelligence Graduate School Program at Korea University) and (No. RS-2024-00457882, National AI Research Lab Project).
\bibliography{aaai2026}



\setcounter{equation}{0}

\newpage
\begin{center}
    {\LARGE \bfseries Appendix of HyFI: \\[0.5em]
     Hyperbolic Feature Interpolation for Brain-Vision Alignment \par}
    \vspace{1.5em}
\end{center}

\setcounter{equation}{0}

\subsection{Image Augmentation}
In this section, we describe Gaussian blur and fovea blur to generate perceptual and semantic images, respectively. We also present our parameter search strategy for identifying the optimal augmentation configuration.

\subsubsection{Gaussian Blur} Gaussian blur removes high-frequency details such as textures and sharp edges, while preserving coarse visual structures like color and shape.
Formally, the Gaussian-blurred image \( x_{\text{blur}} \) is computed as:
\begin{equation}
x_{\text{blur}}(i, j) = \sum_{m=-k}^{k} \sum_{n=-k}^{k} x(i - m, j - n) \cdot G(m, n),
\end{equation}
where \( x(i,j) \) is the pixel value of input image, $r = 2k + 1$ represents the radius of the Gaussian kernel. The Gaussian kernel defined as:
\begin{equation}
G(m,n) = \frac{1}{2\pi\sigma^2} \exp\left(-\frac{m^2 + n^2}{2\sigma^2}\right),
\end{equation}
with $ \sigma$ controlling the strength of the blur.

\subsubsection{Fovea Blur} To simulate human vision, we adopt the fovea blur method described in prior work \cite{wu2025bridging}. 

\begin{equation}
    \tilde{x}_v = \delta \cdot x + (1 - \delta) \cdot x_{\text{blur}},
\end{equation}
where, $\delta$ is the blending factor.
To simulate the fovea effect, the blending weight \( \delta(i,j) \) is defined as a function of the distance from the center:
\begin{equation}
    \delta(i,j) = \exp\left(-\frac{\lambda \cdot d(i,j)}{L}\right),
\end{equation}
where $ d(i,j)$ denotes the Euclidean distance between pixel $ (i,j) $ and the fovea, $ L $ is the maximum possible distance in the image, and $\lambda$ is a hyperparameter that controls the rate of decay.

\begin{figure}[h]
    \centering
    \includegraphics[width=1\linewidth]{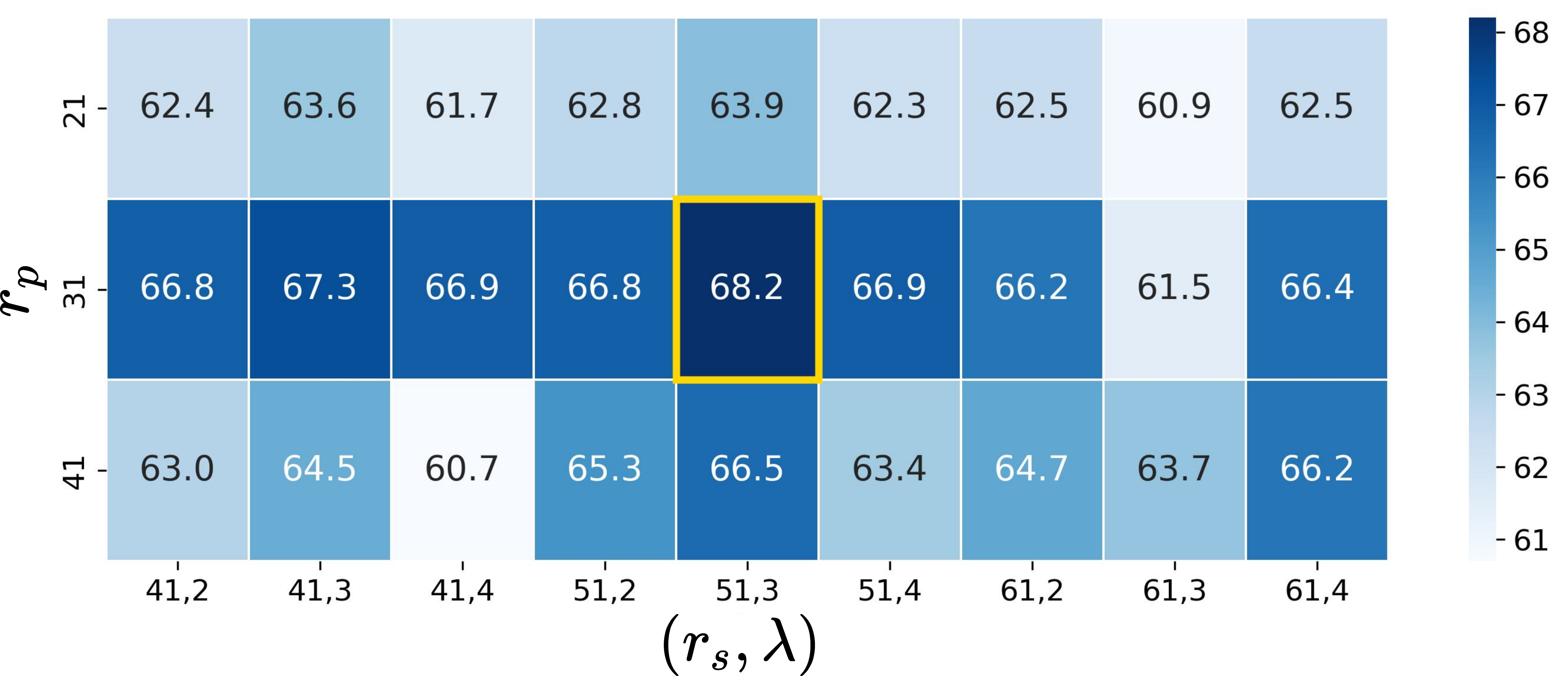}
\caption{Top-1 accuracy heatmap from a parameter search over perceptual and semantic visual augmentation configurations.}

\label{aug_val}
\end{figure}

To investigate the effect of augmentation, we systematically varied the augmentation parameters. Let $r_p$ and $r_s$ denote the Gaussian kernel sizes for perceptual and semantic image generation, respectively. The best Top-1 accuracy on image retrieval task was achieved when $r_p = 31$, $r_s = 51$, and $\lambda = 3$, as show in Fig. \ref{aug_val}.

\subsection{Geodesic Interpolation in Lorentz Model}

\subsubsection{Definition of the Exponential Map}
Following the geodesic definition introduced in the main text, a geodesic $\gamma(t)$ starting at $\mathbf{p} \in \Lspace$ with an initial tangent vector $\mathbf{v}_0 \in T_{\mathbf{p}} \Lspace$ can be parameterized as:
\begin{equation}
   \gamma(t) = \cosh\left( t\sqrt{\kappa} \|\mathbf{v}_0\|_{\mathbb{L}} \right) \mathbf{p} + \frac{\sinh\left( t\sqrt{\kappa} \|\mathbf{v}_0\|_{\mathbb{L}} \right)}{\sqrt{\kappa} \|\mathbf{v}_0\|_{\mathbb{L}}} \mathbf{v}_0.
\end{equation}
For notational convenience, let us define $\beta_0 = \sqrt{\kappa}\norm{\mathbf{v}_0}_\mathbb{L}$. Then, for $t \ge 0$, $\sqrt{\kappa}\norm{t\mathbf{v}_0}_\mathbb{L} = t\beta_0$.
Substituting $t\beta_0$ into the exponential map:
\begin{align}
    \gamma(t) 
    &= \cosh(t\beta_0)\mathbf{p} + \frac{\sinh(t\beta_0)}{\beta_0}\mathbf{v}_0. \label{eq:geodesic_via_expmap_derive}
\end{align}
This formula describes the geodesic extending from $\mathbf{p}$ in the direction of $\mathbf{v}_0$.

\subsubsection{Reformulation Geodesic}
To find the geodesic connecting two points $\mathbf{p}$ and $\mathbf{q}$ in $\Lspace$, we use the logarithm map $\log^\kappa_{\mathbf{p}}(\cdot)$.
We define the point $\mathbf{q}$ as the exponential map of a tangent vector $\mathbf{v}_0$ at $\mathbf{p}$ (i.e.,
$\mathbf{q} = \exp^\kappa_{\mathbf{p}}(\mathbf{v}_0)$).

From Eq.(\ref{eq:geodesic_via_expmap_derive}) with $t=1$, we have $\mathbf{q} = \cosh(\beta_0)\mathbf{p} + \frac{\sinh(\beta_0)}{\beta_0}\mathbf{v}_0$.
Taking the Lorentzian inner product of both sides with $\mathbf{p}$:
\begin{align}
    \langle \mathbf{p}, \mathbf{q} \rangle_{\mathbb{L}} &= \langle \mathbf{p}, \cosh(\beta_0)\mathbf{p} + \frac{\sinh(\beta_0)}{\beta_0}\mathbf{v}_0 \rangle_{\mathbb{L}} \\
    &= \cosh(\beta_0)\langle \mathbf{p}, \mathbf{p} \rangle_{\mathbb{L}} + \frac{\sinh(\beta_0)}{\beta_0}\langle \mathbf{p}, \mathbf{v}_0 \rangle_{\mathbb{L}}
\end{align}
By the Lorentzian constraint, $\langle \mathbf{p}, \mathbf{p} \rangle_{\mathbb{L}} = -1/\kappa$ and $\langle \mathbf{p}, \mathbf{v}_0 \rangle_{\mathbb{L}} = 0$ (since $\mathbf{v}_0 \in T_{\mathbf{p}}\Lspace$). Thus:
\begin{align}
    \langle \mathbf{p}, \mathbf{q} \rangle_{\mathbb{L}} = -\frac{1}{\kappa}\cosh(\beta_0)
\end{align}
From the geodesic distance definition in the main text, $d_{\mathbb{L}}(\mathbf{p}, \mathbf{q}) = \sqrt{1/\kappa} \cdot \cosh^{-1} \left( -\kappa \langle \mathbf{p}, \mathbf{q} \rangle_{\mathbb{L}} \right)$.
Let $\beta = \sqrt{\kappa} \cdot d_{\mathbb{L}}(\mathbf{p}, \mathbf{q})$. Then, $\cosh(\beta) = -\kappa \cdot \langle \mathbf{p}, \mathbf{q} \rangle_{\mathbb{L}}$.
Comparing the two expressions for $\cosh(\beta_0)$ and $\cosh(\alpha)$, we find $\beta_0 = \beta$.
Thus, the parameter $\beta_0$ associated with the tangent vector $\mathbf{v}_0 = \log_{\mathbf{p}}(\mathbf{q})$ is precisely $\beta = \sqrt{\kappa}\cdot d_{\mathbb{L}}(\mathbf{p}, \mathbf{q})$.

Now, we solve for $\mathbf{v}_0$ from $\mathbf{q} = \cosh(\beta)\mathbf{p} + \frac{\sinh(\beta)}{\beta}\mathbf{v}_0$:
\begin{align}
    \mathbf{q} - \cosh(\beta)\mathbf{p} &= \frac{\sinh(\beta)}{\beta}\mathbf{v}_0 \\
    \mathbf{v}_0 &= \frac{\beta}{\sinh(\beta)} (\mathbf{q} - \cosh(\beta)\mathbf{p})
\end{align}
This vector $\mathbf{v}_0$ is exactly what is given by the logarithmic map $\log^\kappa_{\mathbf{p}}(\mathbf{q})$ in the main text.

Finally, substituting this $\mathbf{v}_0$ back into the geodesic definition Eq.(\ref{eq:geodesic_via_expmap_derive}):
\begin{align}
    \gamma(t) &= \cosh(t\beta)\mathbf{p} + \frac{\sinh(t\beta)}{\beta} \mathbf{v}_0 \\
    &= \cosh(t\beta)\mathbf{p} + \frac{\sinh(t\beta)}{\beta} \left( \frac{\beta}{\sinh(\beta)} (\mathbf{q} - \cosh(\beta)\mathbf{p}) \right) \\
    &= \cosh(t\beta)\mathbf{p} + \frac{\sinh(t\beta)}{\sinh(\beta)}\mathbf{q} - \frac{\sinh(t\beta)\cosh(\beta)}{\sinh(\beta)}\mathbf{p} \\
    &= \left( \cosh(t\beta) - \frac{\sinh(t\beta)\cosh(\beta)}{\sinh(\beta)} \right)\mathbf{p} + \frac{\sinh(t\beta)}{\sinh(\beta)}\mathbf{q} 
    \label{eq:early_geodesic}
\end{align}
Using the hyperbolic trigonometric identity $\sinh(A-B) = \sinh(A)\cosh(B) - \cosh(A)\sinh(B)$, the coefficient in front of $\mathbf{p}$ in Eq.~(\ref{eq:early_geodesic}) can be rewritten as:
\begin{align}
    &\cosh(t\beta) - \frac{\sinh(t\beta)\cosh(\beta)}{\sinh(\beta)} \\
    &= \frac{\cosh(t\beta)\sinh(\beta) - \sinh(t\beta)\cosh(\beta)}{\sinh(\beta)} \\
    &= \frac{-\sinh(t\beta - \beta)}{\sinh(\beta)} = \frac{\sinh(\beta - t\beta)}{\sinh(\beta)} = \frac{\sinh((1-t)\beta)}{\sinh(\beta)}
\end{align}
Thus, the geodesic from $\mathbf{p}$ to $\mathbf{q}$ is expressed in terms of hyperbolic sine functions:
\begin{align}
    \gamma(t) = \frac{\sinh((1-t)\beta)}{\sinh(\beta)}\mathbf{p} + \frac{\sinh(t\beta)}{\sinh(\beta)}\mathbf{q}, \quad t \in [0,1]. \label{eq:geodesic_final_form_reconfirmed_2}
\end{align}

\subsection{Convexity of Geodesics in the Hyperbolic Space}

\subsubsection{Problem Setup}
Let $\mathbf{p}, \mathbf{q} \in \Lspace$ be two points on the hyperboloid model separated by a hyperbolic distance $D > 0$, and define $\beta = \sqrt{\kappa}\cdot D$.

Let $\tilde{\mathbf{p}}$ and $\tilde{\mathbf{q}}$ be the spatial components of $\mathbf{p}$ and $\mathbf{q}$, respectively, belonging to $\Rspace$. The spatial part of the geodesic interpolation is:
\begin{align}
    \gamma_{\text{spatial}}(t) &= a\tilde{\mathbf{p}} + b\tilde{\mathbf{q}}, \nonumber\\
    \text{where } a = \frac{\sinh((1-t)\beta)}{\sinh(\beta)}& \text{ and } b = \frac{\sinh(t\beta)}{\sinh(\beta)}.
\end{align}
In contrast, the straight-line (Euclidean) interpolation of the spatial coordinates is:
\begin{align}
    M_{\text{spatial}}(t) = (1-t)\tilde{\mathbf{p}} + t\tilde{\mathbf{q}}. \label{eq:euclidean_again_3}
\end{align}
Our goal is to prove that for every $t \in (0,1)$, the hyperbolic interpolation lies strictly closer to the origin in $\Rspace$ (with the standard Euclidean norm $\norm{\cdot}$) than the Euclidean interpolation, unless $\tilde{\mathbf{p}}$ and $\tilde{\mathbf{q}}$ are collinear with the origin. That is, we aim to show:
\begin{align}
    \norm{\gamma_{\text{spatial}}(t)} < \norm{M_{\text{spatial}}(t)}.\label{eq:goal_again_3}
\end{align}

\subsubsection{Main Proof}

The proof proceeds in three steps. First, we establish key properties of the coefficients $a$ and $b$ relative to their Euclidean counterparts. Second, we establish their sum property. Finally, we use these properties to compare the norms.

\subsubsection{Key Inequalities for Coefficients $a$ and $b$}

\paragraph{Lemma 1 (Coefficient Inequality):}
For $\beta > 0$ and $t \in (0,1)$, the coefficients $a$ and $b$ satisfy:
\begin{align}
    a = \frac{\sinh((1-t)\beta)}{\sinh(\beta)} &< (1-t) \\
    b = \frac{\sinh(t\beta)}{\sinh(\beta)} &< t
\end{align}
\paragraph{Proof}
Consider the function $f(x) = \frac{\sinh(x)}{x}$ for $x > 0$. Its derivative is $f'(x) = \frac{x\cosh(x) - \sinh(x)}{x^2}$.
Let $g(x) = x\cosh(x) - \sinh(x)$. Its derivative is $g'(x) = \cosh(x) + x\sinh(x) - \cosh(x) = x\sinh(x)$.
For $x > 0$, $g'(x) > 0$, so $g(x)$ is strictly increasing. Since $g(0) = 0$, it follows that $g(x) > 0$ for all $x > 0$.
Therefore, $f'(x) > 0$ for $x > 0$, implying that $f(x) = \frac{\sinh(x)}{x}$ is a \textbf{strictly increasing function} for $x > 0$.

Now, we apply this property to $a$ and $b$:
For $a$: Since $t \in (0,1)$ and $\beta > 0$, we have $0 < (1-t)\beta < \beta$.
Because $f(x)$ is strictly increasing:
\begin{align}
    f((1-t)\beta) &< f(\beta) \\
    \frac{\sinh((1-t)\beta)}{(1-t)\beta} &< \frac{\sinh(\beta)}{\beta}
\end{align}
Multiplying both sides by $(1-t)$ (which is positive):
\begin{align}
    \frac{\sinh((1-t)\beta)}{\sinh(\beta)} < (1-t)
\end{align}
So, $a < (1-t)$.

\noindent Similarly, for $b$: Since $0 < t\beta < \beta$:
\begin{align}
    f(t\beta) &< f(\beta) \\
    \frac{\sinh(t\beta)}{t\beta} &< \frac{\sinh(\beta)}{\beta}
\end{align}
Multiplying both sides by $t$ (which is positive):
\begin{align}
    \frac{\sinh(t\beta)}{\sinh(\beta)} < t
\end{align}
So, $b < t$.
$\blacksquare$

\subsubsection{Convexity and Sum of Coefficients}

Let us analyze the sum of the coefficients, $S(t) = a+b$.
\begin{align}
    S(t) = \frac{\sinh((1-t)\beta) + \sinh(t\beta)}{\sinh(\beta)}.
\end{align}
We will show that $S(t)$ is a strictly convex function for $t \in (0,1)$. The second derivative of $S(t)$ with respect to $t$ is:
\begin{align}
    S'(t) &= \frac{-\beta\cosh((1-t)\beta) + \beta\cosh(t\beta)}{\sinh(\beta)}, \\
    S''(t) &= \frac{\beta^2\sinh((1-t)\beta) + \beta^2\sinh(t\beta)}{\sinh(\beta)}.
\end{align}
Since $\beta > 0$ and $t \in (0,1)$, both $(1-t)\beta$ and $t\beta$ are positive. Therefore, $\sinh((1-t)\beta) > 0$ and $\sinh(t\beta) > 0$, which implies $S''(t) > 0$.

Because $S(t)$ is strictly convex, for any $t \in (0,1)$, its value is strictly less than the value on the line segment connecting its endpoints $S(0)$ and $S(1)$.
We evaluate the function at the endpoints:
\begin{align}
    S(0) &= \frac{\sinh(\beta) + \sinh(0)}{\sinh(\beta)} = 1, \\
    S(1) &= \frac{\sinh(0) + \sinh(\beta)}{\sinh(\beta)} = 1.
\end{align}
By strict convexity, for all $t \in (0,1)$, we have $S(t) < \max(S(0), S(1)) = 1$. This means:
\begin{align}
    a+b < 1. \label{eq:sum_less_than_one_final_2}
\end{align}
This result also follows directly from Lemma 1, since $a < (1-t)$ and $b < t$ implies $a+b < (1-t)+t = 1$. However, the convexity argument provides an elegant alternative.

\subsubsection{Comparison of Norms}

We want to compare $\norm{\gamma_{\text{spatial}}(t)}$ and $\norm{M_{\text{spatial}}(t)}$. Let $\mathbf{P} = \tilde{\mathbf{p}}$ and $\mathbf{Q} = \tilde{\mathbf{q}}$ for simplicity.
Let's consider the Euclidean triangle $\mathcal{T}$ with vertices at the origin $\mathbf{0}$, $\mathbf{P}$, and $\mathbf{Q}$.
\begin{itemize}
    \item The geodesic interpolation $\gamma_{\text{spatial}}(t) = a\mathbf{P} + b\mathbf{Q}$. From Lemma 1, $a > 0$ and $b > 0$. From the sum of coefficients property, $a+b < 1$. This means $\gamma_{\text{spatial}}(t)$ is a strict convex combination of $\mathbf{0}$, $\mathbf{P}$, and $\mathbf{Q}$ (since the coefficient for $\mathbf{0}$ is $1-(a+b)$, which is strictly positive). Therefore, $\gamma_{\text{spatial}}(t)$ lies strictly in the \textbf{open interior} of the triangle $\mathcal{T}$.
    \item The Euclidean interpolation $M_{\text{spatial}}(t) = (1-t)\mathbf{P} + t\mathbf{Q}$. For $t \in (0,1)$, $(1-t) > 0$, $t > 0$, and their sum is $(1-t)+t=1$. Therefore, $M_{\text{spatial}}(t)$ lies strictly on the \textbf{line segment} connecting $\mathbf{P}$ and $\mathbf{Q}$. This segment is the side of the triangle $\mathcal{T}$ that is opposite to the origin $\mathbf{0}$.
\end{itemize}

Now, consider the case where $\mathbf{P}$ and $\mathbf{Q}$ are non-collinear with the origin. In this scenario, the triangle $\mathcal{T}$ is non-degenerate. A fundamental geometric property of Euclidean space states that for a non-degenerate triangle with one vertex at the origin, any point in the open interior of the triangle is strictly closer to the origin than any point on the opposite side (the side that does not include the origin). This holds true unless the opposite side itself passes through the origin. Since we are in the non-collinear case, the segment connecting $\mathbf{P}$ and $\mathbf{Q}$ does not pass through the origin.

This geometric property thus proves our initial claim from Eq. (22):
\begin{center}
    $||\gamma_{\text{spatial}}(t)|| < ||M_{\text{spatial}}(t)||$.
\end{center}
This inequality is strict for all $t \in (0,1)$, unless $\tilde{\mathbf{p}}$ and $\tilde{\mathbf{q}}$ are collinear with the origin.

\subsubsection{Degenerate Collinear Case}
If $\tilde{\mathbf{p}}$ and $\tilde{\mathbf{q}}$ are collinear with the origin (i.e., $\tilde{\mathbf{q}} = c\tilde{\mathbf{p}}$ for some scalar $c$), the "triangle" formed by $\mathbf{0}$, $\tilde{\mathbf{p}}$, and $\tilde{\mathbf{q}}$ degenerates into a line segment passing through the origin. In this radial case, both $\gamma_{\text{spatial}}(t)$ and $M_{\text{spatial}}(t)$ lie on this line segment.
The spatial components become:
\begin{align}
    \gamma_{\text{spatial}}(t) &= (a + bc)\tilde{\mathbf{p}} \\
    M_{\text{spatial}}(t) &= ((1-t) + tc)\tilde{\mathbf{p}}
\end{align}
The proof still holds true: from the coefficient inequalities ($a < (1-t)$ and $b < t$), it follows that $\norm{a + bc} < \norm{(1-t)+tc}$ because the coefficients $a$ and $b$ are strictly smaller than their Euclidean counterparts $(1-t)$ and $t$. This ensures that $\gamma_{\text{spatial}}(t)$ remains strictly closer to the origin than $M_{\text{spatial}}(t)$ for $t \in (0,1)$, unless $t=0$ or $t=1$. The only edge case where the norms might be equal is if the points $\tilde{\mathbf{p}}$ and $\tilde{\mathbf{q}}$ are the zero vector, which corresponds to the hyperbolic points being at the origin of their respective tangent spaces, but this is a specific degenerate scenario generally excluded by the problem's context of separated points.

\subsubsection{Relationship Between Spatial Norm and Time Component}
As defined in Eq. (1) in the main text, any point $\mathbf{p} = (\mathbf{p}_0, \tilde{\mathbf{p}}) \in \mathbb{L}^n$ must satisfy the constraint $\langle \mathbf{p}, \mathbf{p} \rangle_{\mathbb{L}} = -1/\kappa$. Using the definition of the Lorentzian inner product from Eq. (2) in the main text, this constraint can be expanded for a single point $\mathbf{p}$:
\begin{align}
    -\mathbf{p}_0^2 + \langle \tilde{\mathbf{p}}, \tilde{\mathbf{p}} \rangle_E = -1/\kappa.
\end{align}
Here, $\langle \tilde{\mathbf{p}}, \tilde{\mathbf{p}} \rangle_E$ is the standard Euclidean dot product, which is equivalent to the squared Euclidean norm $||\tilde{\mathbf{p}}||^2$ of the spatial component $\tilde{\mathbf{p}} \in \mathbb{R}^n$. By rearranging the equation to solve for the time component $\mathbf{p}_0$ (given $\mathbf{p}_0 > 0$), we obtain a direct relationship:
\begin{equation}
    \mathbf{p}_0 = \sqrt{||\tilde{\mathbf{p}}||^2 + 1/\kappa}.
\end{equation}
This equation mathematically demonstrates that the time component $\mathbf{p}_0$ is a monotonically increasing function of the spatial component's norm $||\tilde{\mathbf{p}}||$. Therefore, a point being ``closer to the spatial origin'' (i.e., having a smaller $||\tilde{\mathbf{p}}||$) is equivalent to its time component $\mathbf{p}_0$ being smaller. This brings it closer to the Lorentz manifold's origin point $\mathbf{O} = (\frac{1}{\sqrt{\kappa}}, 0, \dots, 0)^T \in \mathbb{L}^n$, which is the point with the minimum possible time component. The conclusion of our proof that $||\gamma_{\text{spatial}}(t)|| < ||M_{\text{spatial}}(t)||$ thus directly implies that the hyperbolic geodesic interpolation lies closer to the manifold's origin than its Euclidean counterpart.

\subsection{Dataset}
To study visual brain decoding, we utilized paired datasets comprising brain signals and corresponding visual stimuli.  Specifically, we used the widely adopted THINGS-EEG and THINGS-MEG datasets. Both datasets were collected under the Rapid Serial Visual Presentation (RSVP) paradigm, which presents images in rapid succession while recording time-resolved neural responses.

\subsubsection{THINGS-EEG} The THINGS-EEG dataset \cite{gifford2022large} comprises EEG–image pairs collected from 10 subjects. EEG signals were recorded using a 64-channel system at a sampling rate of 1000 Hz and filtered to the range [0.1, 100] Hz. The training set consists of 1,654 object concepts, each associated with 10 distinct images, and each image repeated 4 times per subject (total 1,654 $\times$ 10 $\times$ 4 samples). The test set includes 200 concepts, each represented by a single image repeated 80 times per subject (total 200 $\times$ 80 samples).

To ensure fair comparison with prior work, we follow the preprocessing protocol established in~\cite{songdecoding, wu2025bridging}. The raw EEG signals are downsampled to 250 Hz, and 17 channels located over occipital and parietal regions—areas associated with visual processing—are selected. All repetitions for each image are averaged to enhance the signal-to-noise ratio (SNR), resulting in 16,540 training samples and 200 test samples per subject.

\subsubsection{THINGS-EEG} The THINGS-MEG dataset \cite{hebart2023things} contains MEG–image pairs acquired from four participants using a 271-channel. During each trial, a visual stimulus was presented for 500 ms, followed by an inter-stimulus interval consisting of a blank screen lasting 1000 ± 200 ms. The training set includes 1,854 object concepts, each associated with 12 unique images presented once per subject. The test set comprises 200 novel concepts, each represented by a single image repeated 12 times.

We follow the pre-proesccing  pipeline described in~\cite{songdecoding, wu2025bridging}. MEG signals are bandpass filtered between 0.1 and 100 Hz and subsequently downsampled to 200 Hz. To enhance signal quality, all repetitions are averaged. Furthermore, we exclude the 200 concepts from the training set, consistent with prior experimental protocols.

\begin{table}[t]
\footnotesize
\setlength{\tabcolsep}{2.5pt}
\centering
\begin{tabular}{lcccccccc}
\toprule
Model & Params & Emb dim & \multicolumn{2}{c}{Semantic} & \multicolumn{2}{c}{Perceptual} \\

& & & Norm & Std & Norm & Std \\
\midrule
\midrule
RN50         & 38.32 M   & 1024 & 233  & 0.05 & 277  & 0.06 \\
RN101        & 56.26 M   & 512  & 282  & 0.09 & 325  & 0.11 \\
ViT/B-16     & 86.19 M   & 512  & 1409 & 0.48 & 1404 & 0.48 \\
ViT/B-32     & 87.85 M   & 512  & 1447 & 0.49 & 1473 & 0.51 \\
ViT/L-14     & 303.97 M  & 768  & 2399 & 0.67 & 2535 & 0.71 \\
ViT/H-14     & 632.08 M  & 1024 & 2835 & 0.68 & 2587 & 0.62 \\
\midrule
MERU (ViT-S) & 21.66 M   & 512  & 160  & 0.05 & 162  & 0.05 \\
MERU (ViT-B) & 89.79 M   & 512  & 320  & 0.11 & 319  & 0.11 \\
MERU (ViT-L) & 303.30M   & 512  & 414  & 0.14 & 411  & 0.14 \\
\midrule
HyCoClip (ViT-S)  & 21.66 M   & 512  & 80   & 0.03 & 79   & 0.03 \\
HyCoClip (ViT-B)  & 89.79 M  & 512  & 118  & 0.04 & 118  & 0.04 \\
\midrule
\bottomrule
\end{tabular}
\caption{Comparison of vision encoders and feature statistics.}
\label{info_visual}
\end{table}

\subsection{Baseline Descriptions}
\begin{itemize}

    \item \textbf{BraVL} \cite{du2023decoding} introduces a Mixture-of-Experts (MoE) framework that jointly models brain, visual, and linguistic modalities for neural decoding. The model leverages multiple expert pathways to capture complementary information across modalities.
    \item \textbf{NICE} \cite{songdecoding} adopts a self-supervised contrastive learning framework, incorporating dual spatial attention modules to enhance EEG feature representations.
    \item \textbf{ATM} \cite{li2024visual} proposes a dedicated EEG encoder named Adaptive Thinking Mapper, which integrates positional encoding with temporal-spatial EEG features to better model brain dynamics.
    \item \textbf{Cog-cap}~\cite{zhang2025cognitioncapturer} presents a multi-modal architecture that captures cross-modal information—such as text, depth, and image representations—by processing EEG signals through multiple expert encoders.
    \item \textbf{UBP} \cite{wu2025bridging} is a recent state-of-the-art approach that introduces uncertainty-aware fovea blur to enhance the alignment between brain signals and visual features.
\end{itemize}

\begin{table}[t]
\centering
\begin{tabular}{llcc}
\toprule
Semantic & Perceptual & Top-1 ACC & Top-5 ACC \\
\midrule
\midrule
\multirow{5}{*}{Original}
  & Fovea blur      & 53.8 & 83.7 \\
  & SAM             & 56.6 & 85.4 \\
  & Gaussian noise  & 61.4 & 89.8 \\
  & Low resolution  & 63.5 & 89.6 \\
  & Gaussian blur   & 65.3 & 91.4 \\
\midrule
\multirow{4}{*}{Fovea blur}
  & SAM             & 60.5 & 88.5 \\
  & Gaussian noise  & 66.3 & 92.7 \\
  & Low resolution  & 65.6 & 91.2 \\
  & Gaussian blur   & \textbf{68.2} & \textbf{91.9} \\
\midrule
\bottomrule
\end{tabular}
\caption{Image retrieval accuracy ($\%$)  with various semantic and perceptual augmentations.}
\label{aug_varation_results}
\end{table}

\begin{figure}[h]
    \centering
    \includegraphics[width=0.9\linewidth]{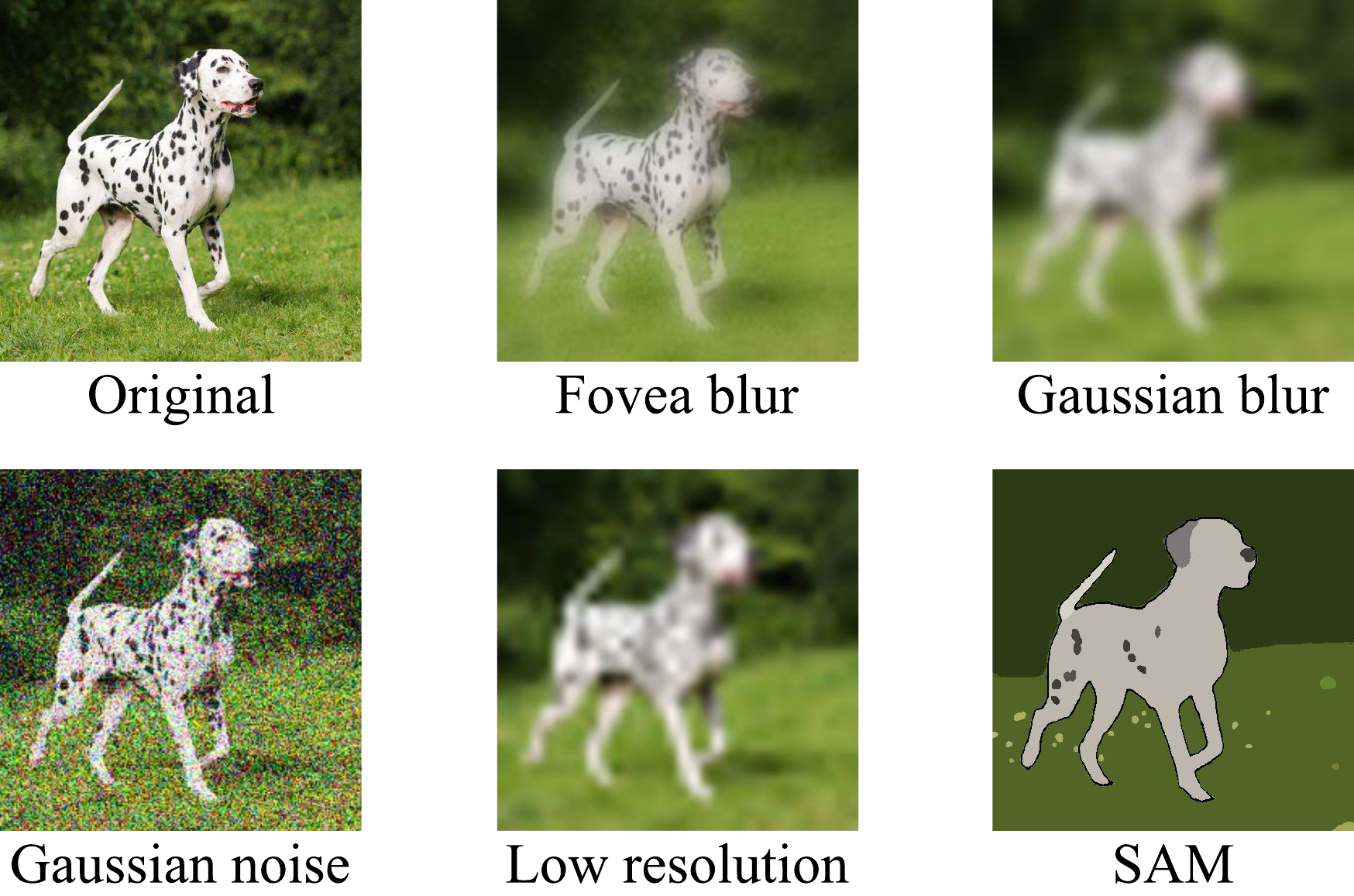}
\caption{Examples of different augmentations. SAM refer to SAM-based augmentation that segments the image and fills each region with its mean color. SAM refers to an augmentation that uses the SAM to segment the image and fills each region with its mean color.
}
\label{aug_variation}
\end{figure}

\subsection{Ablation Study of Augmentation}
To obtain effective perceptual and semantic features, we applied various augmentations to a diverse set of images to obtain more effective perceptual and semantic visual features. Specifically, we consider fovea blur, Gaussian blur, Gaussian noise, low-resolution, and a segment anything model(SAM)-based augmentation that segments the image and fills each region with its mean color. Examples of these augmentations are shown in Fig.~\ref{aug_variation}. 

We find that applying fovea blur to semantic images yields better performance, which aligns with prior findings~\cite{wu2025bridging} suggesting that simulating human visual attention through augmentation can be beneficial. Among the perceptual augmentations, Gaussian blur consistently outperforms other augmentations. Based on these observations, we use fovea blur and Gaussian blur for semantic and perceptual images, respectively.

\subsection{Interpolation Coefficient}
To adaptively integrate perceptual and semantic features, we introduce an interpolation coefficient $t$. In our main text, $t$ is computed from the semantic feature $\mathbf{x}_v^{s}$ as follows:
\begin{equation}
t = \sigma(W_t f_v(\mathbf{x}_v^{s})),
\end{equation}
where $\sigma$ denotes the sigmoid function, $W_t \in \mathbb{R}^{1\times d}$ is a learnable weight vector, $d$ is the feature dimension, and $f_v$ is the pre-trained visual encoder.

Table~\ref{score_abl} reports the retrieval performance obtained by varying the input feature used to compute the interpolation coefficient $t$. Using EEG features resulted in the lowest performance, while different image-based features showed comparable results with no substantial differences.

\begin{table}[t]
\centering
\begin{tabular}{lcc}
\toprule
Input feature & Top-1 ACC & Top-5 ACC \\
\midrule
EEG       & 54.5     & 84.8               \\
Perceptual      & 67.1               & 91.8      \\
Original  & 67.6               & 91.8      \\
Semantic  & \textbf{68.2}      & \textbf{91.9}      \\
\bottomrule
\end{tabular}
\caption{Retrieval accuracy with different strategies for obtaining the interpolation coefficient $t$.}
\label{score_abl}
\end{table}

\subsection{Feature Visualization}
\begin{figure}[h]
\centering
\includegraphics[width=0.65\linewidth]{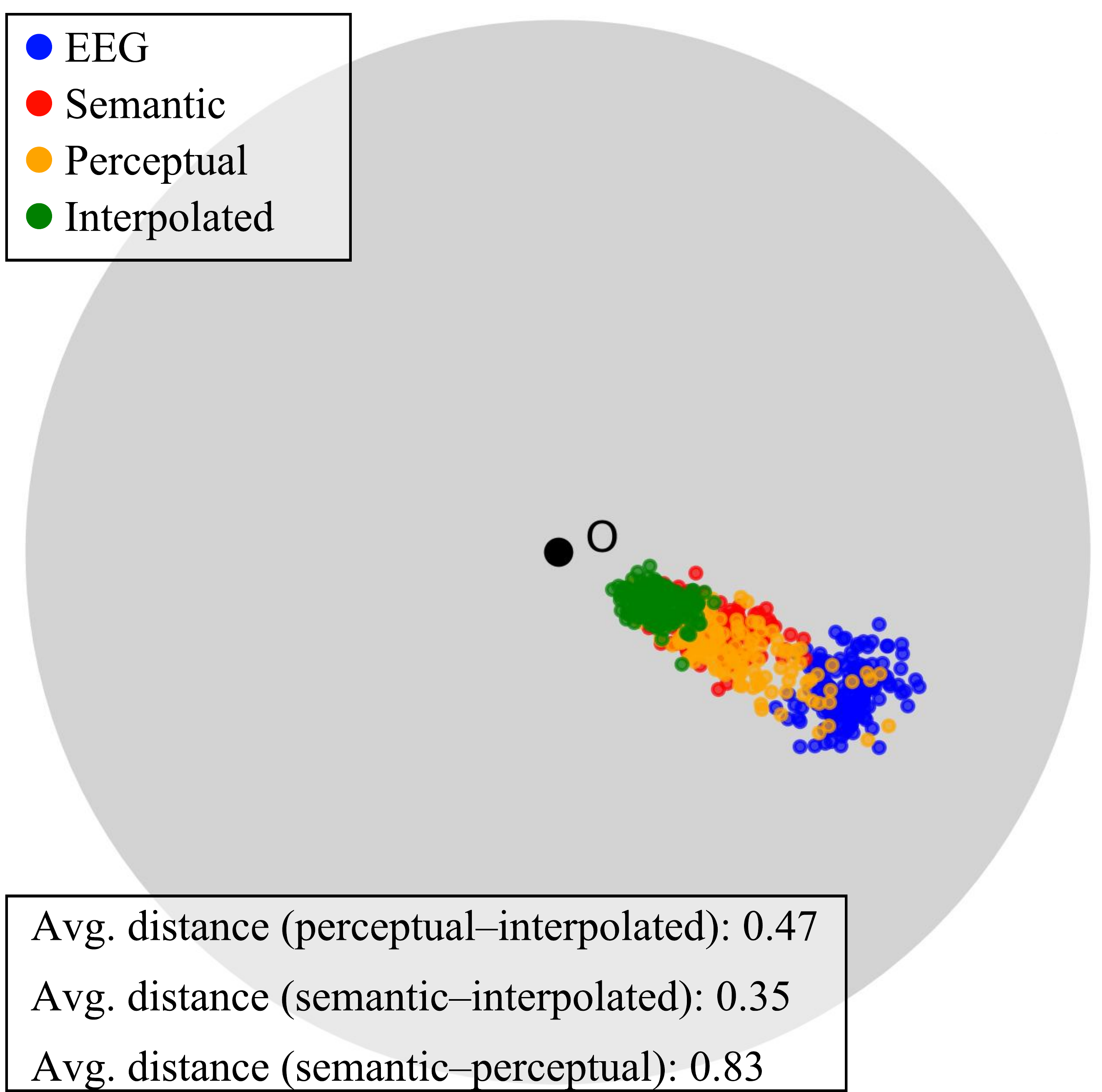}
\caption{Feature distributions visualized on the Poincaré ball using HoronPCA. Interpolated visual features lie between perceptual and semantic features.}
    \label{horonpca}
\end{figure}

To further explore the structure of the learned features, we visualize them on the Poincaré ball using HoronPCA~\cite{chami2021horopca}, as shown in Fig.~\ref{horonpca}. The interpolated visual features lie between the semantic and perceptual visual features and tend to cluster near the origin. Notably, while perceptual and semantic features appear close in the Poincaré ball, their true distance exceeds that to the interpolated feature. This reflects the negative curvature of hyperbolic space, where geodesics through the origin are shorter than lateral paths at constant radius. In contrast, EEG features are embedded far from the origin, naturally reflecting their high variability in the hyperbolic space.

\begin{table}[t]
\centering
\footnotesize
\setlength{\tabcolsep}{1pt}
\begin{tabular}{llcccccccc}
\toprule
\multirow{2}{*}{Method} & \multirow{2}{*}{Model} & \multicolumn{2}{c}{ShallowNet} & \multicolumn{2}{c}{EEGNet} & \multicolumn{2}{c}{TSConv} & \multicolumn{2}{c}{EEGProject} \\
\cmidrule(lr){3-4} \cmidrule(lr){5-6} \cmidrule(lr){7-8} \cmidrule(lr){9-10}
& & Base & Ours & Base & Ours & Base & Ours & Base & Ours \\
\midrule
\midrule
\multirow{6}{*}{CLIP} 
& RN50       & 36.6 & \textbf{50.5} & 36.2 & \textbf{51.6} & 41.1 & \textbf{57.1} & 49.4 & \textbf{68.2} \\
& RN101      & 34.0 & \textbf{45.8} & 35.6 & \textbf{46.4} & 35.9 & \textbf{53.4} & 44.4 & \textbf{62.1} \\
& ViT-B/16   & 28.3 & \textbf{36.7} & 30.7 & \textbf{35.1} & 31.4 & \textbf{40.5} & 36.0 & \textbf{41.6} \\
& ViT-B/32   & 30.7 & \textbf{36.8} & 33.7 & \textbf{37.0} & 36.6 & \textbf{40.8} & 42.5 & \textbf{46.9} \\
& ViT-L/14   & 19.8 & \textbf{22.8} & 19.1 & \textbf{27.0} & 20.9 & \textbf{27.4} & 30.1 & \textbf{34.1} \\
& ViT-H/14   & 19.5 & \textbf{28.3} & 28.4 & \textbf{33.1} & 30.9 & \textbf{35.4} & 42.5 & \textbf{43.8} \\
\midrule
\multirow{3}{*}{MERU}
& ViT-S/16      & 31.8 & \textbf{44.9} & 35.4 & \textbf{45.3} & 38.9 & \textbf{40.0} & 43.7 & \textbf{52.4} \\
& ViT-B/16      & 24.0 & \textbf{33.6} & 28.1 & \textbf{39.1} & 31.5 & \textbf{42.2} & 31.7 & \textbf{48.6} \\
& ViT-L/16      & 16.7 & \textbf{28.8} & 21.0 & \textbf{27.8} & 23.8 & \textbf{33.1} & 23.1 & \textbf{30.9} \\
\midrule
\multirow{2}{*}{HyCoCLIP}
& ViT-S/16      & 35.1 & \textbf{47.9} & 34.4 & \textbf{40.6} & 41.3 & \textbf{47.5} & 44.3 & \textbf{51.6} \\
& ViT-B/16      & 27.0 & \textbf{37.4} & 29.1 & \textbf{34.9} & 32.9 & \textbf{40.6} & 35.6 & \textbf{45.6} \\
\midrule
\bottomrule
\end{tabular}
\caption{
Top-1 accuracy (\%) for various combinations of brain encoders and vision models on brain-to-image retrieval in the THING-EEG dataset. Base indicates performance using either contrastive alignment (CLIP) or hyperbolic contrastive alignment (MERU and HyCoCLIP), without our proposed interpolation method.
}
\label{full}
\end{table}

\subsection{Image Retrieval Results}
\begin{figure*}[t!]
\includegraphics[width=\linewidth]{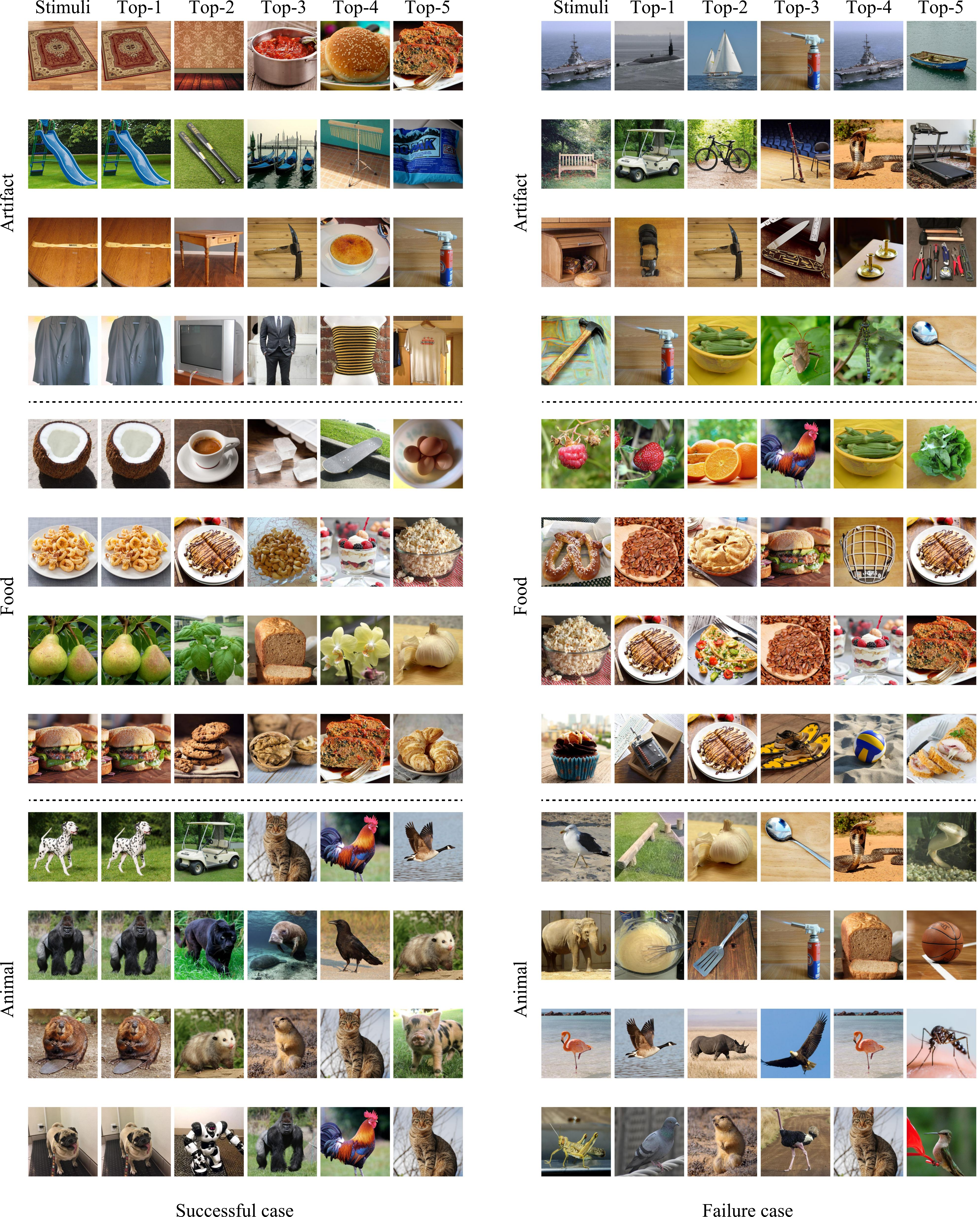}
    \caption{Top-5 image retrieval results on the THINGS-EEG dataset for successful and failure cases.}
    \label{example1}
\end{figure*}

\begin{figure*}[t!]
\includegraphics[width=\linewidth]{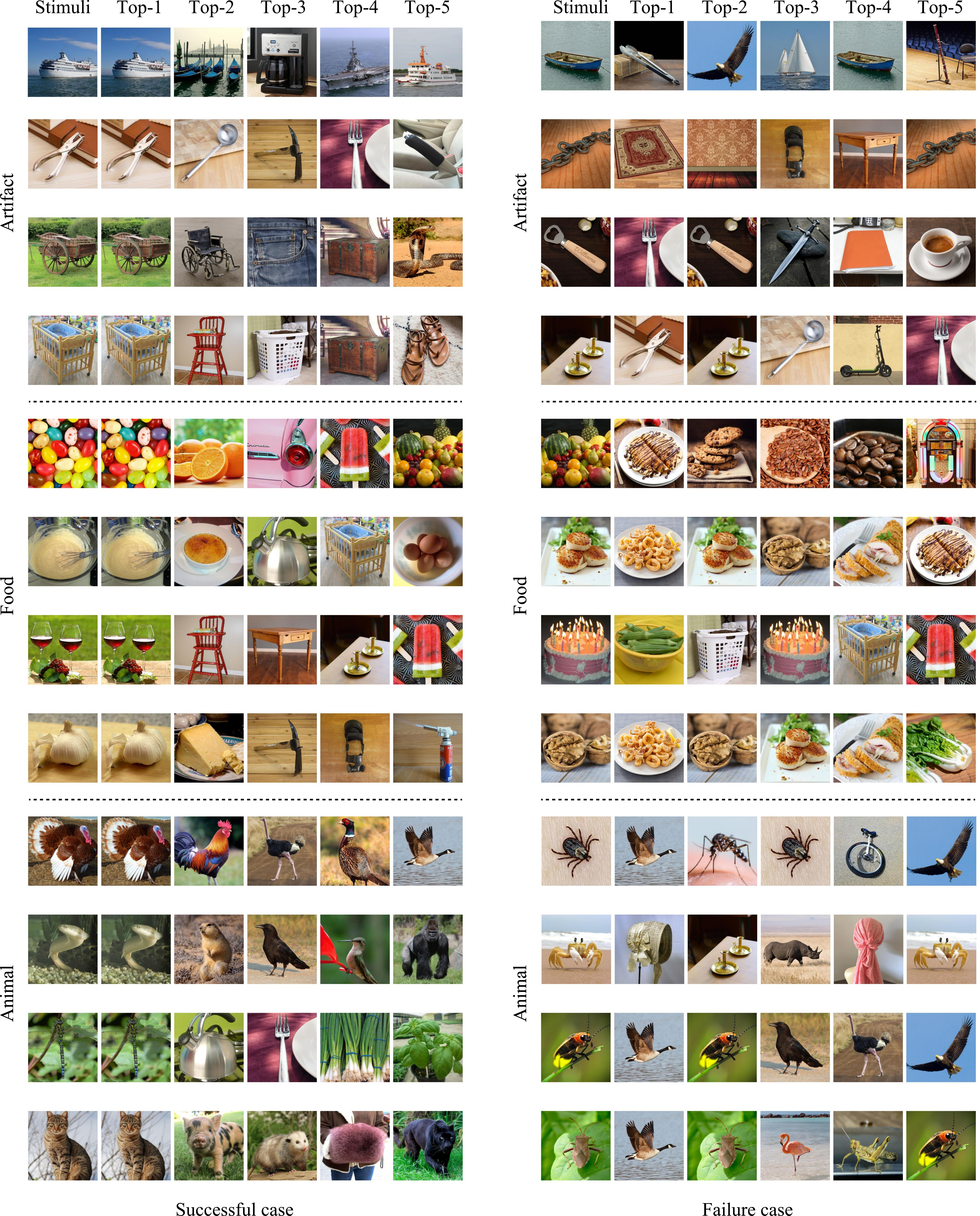}
    \caption{Top-5 image retrieval results on the THINGS-EEG dataset for successful and failure cases.}
    \label{example2}
\end{figure*}

We present the top-5 retrieval results on the THINGS-EEG dataset for both successful and failure cases, categorized into three semantic groups—animals, food, and artifacts—in Fig.~\ref{example1} and Fig.~\ref{example2}. In successful cases, retrieved images align semantically with the stimulus and share perceptual traits like orientation and color. Failure cases rely on superficial cues, such as background color, missing the target concept.

\subsection{Augmentation Effect in Various VLMs}
\begin{figure*}[t!]
\includegraphics[width=\linewidth]{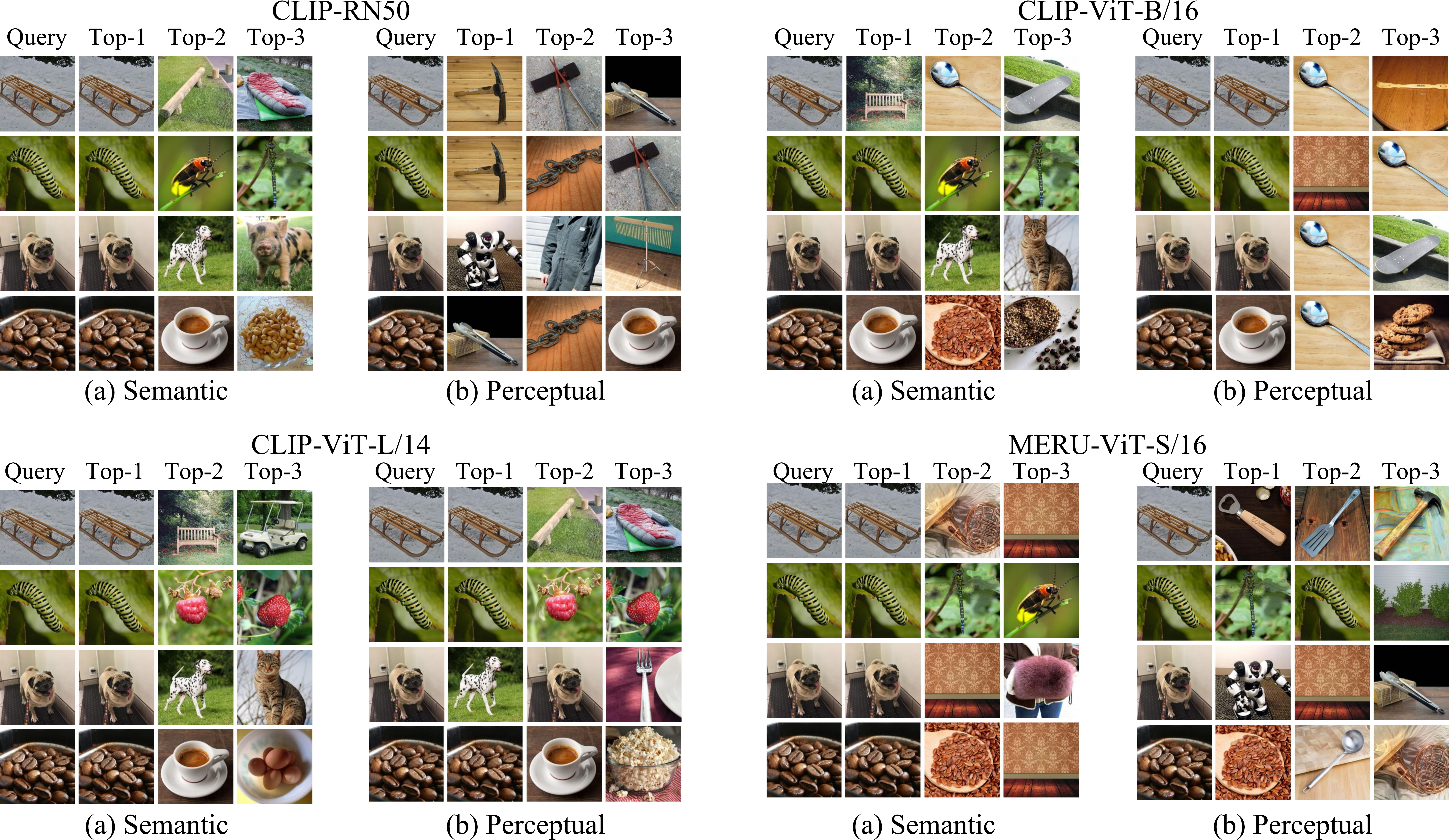}
    \caption{Image retrieval results for various VLMs}
    \label{vlm_aug}
\end{figure*}

We analyze how our image augmentations behave across different vision-language models (VLMs). Specifically, we obtain embeddings of augmented images and retrieve nearby original images in the embedding space to examine which image representations the augmentations are closest to. The results are shown in Fig.~\ref{vlm_aug}. In CLIP-RN50, images with fovea blur predominantly retrieved semantically aligned images, whereas images with Gaussian blur tended to retrieve perceptually similar images, such as those sharing similar orientation or color. However, this effect was less pronounced in other VLMs; in particular, CLIP-ViT-L/14 exhibited minimal sensitivity to perceptual differences. These observations suggest that our method is more effective when applied to ResNet-based CLIP models.

\end{document}